\documentclass{article}

\usepackage[nonatbib, final]{neurips_data_2021}
\usepackage[numbers]{natbib}

\usepackage[utf8]{inputenc} 
\usepackage[T1]{fontenc}    
\usepackage{url}            
\usepackage{booktabs}       
\usepackage{amsfonts}       
\usepackage{nicefrac}       
\usepackage{microtype}      
\usepackage[dvipsnames]{xcolor}
\usepackage{enumitem}
\usepackage{graphicx}
\usepackage{xspace}
\usepackage{soul}
\usepackage{multirow}
\usepackage{bbold}
\usepackage{makecell}
\usepackage{hyperref}       
\usepackage{cleveref}
\usepackage{bm}

\newcommand{\method}{AdvGLUE\xspace}

\newcommand{\m}[1]{{\textcolor{black}{{#1}}}}

\title{Adversarial GLUE: A Multi-Task Benchmark for Robustness Evaluation of Language Models}

\author{%
  \thanks{Equal Contribution}$\,$ Boxin Wang$^1$, $^*$Chejian Xu$^2$, Shuohang Wang$^3$, Zhe Gan$^3$, \\
  \bf Yu Cheng$^3$, Jianfeng Gao$^3$, Ahmed Hassan Awadallah$^3$, Bo Li$^1$ \\
  $^1$University of Illinois at Urbana-Champaign \\
  $^2$Zhejiang University,
  $^3$Microsoft Corporation\\
  \texttt{\small\{boxinw2,lbo\}@illinois.edu, xuchejian@zju.edu.cn}  \\
  \texttt{\small \{shuohang.wang,zhe.gan,yu.cheng,jfgao,hassanam\}@microsoft.com} \\ 
}

\begin{document}

\maketitle

\begin{abstract}
Large-scale pre-trained language models have achieved tremendous success across a wide range of natural language understanding (NLU) tasks, even surpassing human performance. However, recent studies reveal that the robustness of these models can be challenged by carefully crafted textual adversarial examples. While several individual datasets have been proposed to evaluate model robustness,
a principled and comprehensive benchmark is still missing. In this paper, we present Adversarial GLUE (\method), a new multi-task benchmark to quantitatively and thoroughly explore and evaluate the vulnerabilities of modern large-scale language models under various types of adversarial attacks. In particular, we systematically apply 14 textual adversarial attack methods to GLUE tasks to construct AdvGLUE, which is further validated by humans for reliable annotations.
Our findings are summarized as follows.
($i$) Most existing adversarial attack algorithms are prone to generating invalid or ambiguous adversarial examples, with around $90\%$ of them either changing the original semantic meanings or misleading  human annotators as well. Therefore, we perform careful filtering process to curate a high-quality benchmark.
($ii$) All the language models and robust training methods we tested perform poorly on \method, with scores lagging far behind the benign accuracy.
We hope our work will motivate the development of new adversarial attacks that are more stealthy and semantic-preserving, as well as new robust language models against sophisticated adversarial attacks. AdvGLUE is available at \url{https://adversarialglue.github.io}.
\end{abstract}

\section{Introduction}

Pre-trained language models \citep{bert,roberta,Lan2019ALBERTAL,Yang2019XLNetGA,he2020deberta,zhang2019ernie,smart,clark2020electra} have achieved state-of-the-art performance over a wide range of Natural Language Understanding (NLU) tasks \citep{wang2018glue,wang2019superglue,advsquad,advfever,anli}. 
However, recent studies \citep{textfooler,comattack,t3,bertattack,bae} reveal that even these large-scale language models are vulnerable to carefully crafted adversarial examples, which can fool the models to output arbitrarily wrong answers by perturbing input sentences in a human-imperceptible way. 
Real-world systems built upon these vulnerable models can be misled in ways that would have profound security concerns \citep{textbugger,textshield}.

To address this challenge, various methods \citep{smart,freelb,wang2021infobert,alum} have been proposed to improve the adversarial robustness of language models. 
However, the adversary setup considered in these methods lacks a unified standard. 
For example, \citet{smart, alum} mainly evaluate their robustness against human-crafted adversarial datasets \citep{anli,advsquad}, while \citet{wang2021infobert}  evaluate the model robustness  against automatic adversarial attack algorithms \citep{textfooler}.  
The absence of a principled adversarial benchmark makes it difficult to compare the 
robustness across different models and identify the adversarial attacks that most models are  vulnerable to.
This motivates us to build a unified and principled robustness evaluation benchmark for natural language models and hope to help answer the following questions:\textit{ what types of language models are more robust when evaluated on the unified adversarial benchmark? Which adversarial attack algorithms against language models are more effective, transferable, or stealthy to human? How likely can human be fooled by different adversarial attacks?
}

We list out the fundamental principles to create a high-quality robustness evaluation benchmark as follows. 
First, as also pointed out by \citep{criteria}, a reliable benchmark should be accurately and unambiguously annotated by \m{humans}. 
This is especially crucial for the robustness evaluation, as some adversarial examples generated by automatic attack algorithms can fool humans as well \citep{morris-etal-2020-reevaluating}. Given our analysis in \S \ref{sec:attack}, among the generated adversarial data, there are only around $10\%$ adversarial examples that receive at least 4-vote consensus among 5 annotators and align with the original label. 
Thus, additional rounds of human filtering are critical to validate the quality of the generated adversarial attack data. 
Second, a comprehensive robustness evaluation benchmark should cover enough language phenomena and generate a systematic diagnostic report to understand and analyze the vulnerabilities of language models. 
Finally, a robustness evaluation benchmark needs to be challenging and unveil the biases shared across different models.

In this paper, we introduce Adversarial GLUE (\method), a multi-task benchmark for robustness evaluation of language models.
Compared to existing adversarial datasets, there are several contributions that render \method a unique and valuable asset to the community. 
\begin{itemize}[leftmargin=*]
    \item \textbf{Comprehensive Coverage.} We consider textual adversarial attacks from different perspectives and hierarchies, including word-level transformations, sentence-level manipulations, and human-written adversarial examples, so that \method is able to cover as many adversarial linguistic phenomena as possible.
    \item \textbf{Systematic Annotations.} 
    To the best of our knowledge, this is the first work that performs systematic and comprehensive evaluation and annotation over 14 different textual adversarial examples. Concretely, \method adopts  crowd-sourcing to identify high-quality adversarial data for reliable evaluation.
    \item \textbf{General Compatibility.} To obtain comprehensive understanding of the robustness of language models across different NLU tasks, \method covers the widely-used GLUE tasks and creates an adversarial version of the GLUE benchmark to evaluate the robustness of language models.
    \item \textbf{High Transferability and Effectiveness.} \method has high adversarial transferability and can effectively attack a wide range of  state-of-the-art models.  We observe a significant performance drop for models evaluated on \method  compared with their standard accuracy on GLUE leaderboard. For instance, the average GLUE score of ELECTRA(Large)~\cite{clark2020electra} drops from $93.16$ to $41.69$.
\end{itemize}

Our contributions are summarized as follows. ($i$) We propose \method, a principled and comprehensive benchmark  that focuses on  robustness evaluation of language models. ($ii$) During the data construction, we provide a thorough analysis and a fair comparison of existing strong adversarial attack algorithms. ($iii$) We present thorough robustness evaluation for existing state-of-the-art language models and defense methods.
We hope that \method will inspire active research and discussion in the community. More details are available at  \url{https://adversarialglue.github.io}.

\section{Related Work}

Existing robustness evaluation work can be roughly divided into two categories:  \textbf{Evaluation Toolkits} and \textbf{Benchmark Datasets}. 
($i$) Evaluation toolkits, including OpenAttack \citep{openattack}, TextAttack \citep{textattack}, TextFlint \citep{textflint} and Robustness Gym \citep{robustnessgym}, integrate various \textit{ad hoc} input transformations for different tasks and provide programmable APIs to dynamically test model performance. 
However, it is challenging to guarantee the quality of these input transformations. 
For example, as reported in \citep{comattack}, the validity of adversarial transformation can be as low as $65.5\%$, which means that more than one third of the adversarial sentences have wrong labels.
Such a high percentage of annotation errors could lead to an underestimate of model robustness, making it less qualified to serve as an accurate and reliable benchmark \citep{criteria}. 
($ii$) Benchmark datasets for robustness evaluation create challenging testing cases by using human-crafted templates or rules \citep{advfever,checklist,stresstest}, or adopting a human-and-model-in-the-loop manner to write adversarial examples \citep{anli,dynabench,dynabenchqa}. 
While the quality and validity of these adversarial datasets can be well controlled, the scalability and comprehensiveness are limited by the human annotators. 
For example, template-based methods require linguistic experts to carefully construct reasonable rules for specific tasks, and such templates can be barely transferable to other tasks. 
Moreover, human annotators tend to complete the writing tasks through minimal efforts and shortcuts \citep{burghardt2020origins,wall2021left}, which can limit the coverage of various linguistic phenomena.

\begin{table}[t]
\small
    \centering
    \caption{\small \textbf{Statistics  of \method benchmark}. We apply \emph{all} word-level perturbations (C1=\textit{Embedding-similarity}, C2=\textit{Typos}, C3=\textit{Context-aware}, C4=\textit{Knowledge-guided}, and C5=\textit{Compositions}) to the five GLUE tasks. For sentence-level perturbations, we apply \textit{Syntactic-based perturbations} (C6) to the five GLUE tasks. \textit{Distraction-based perturbations} (C7) are applied to four GLUE tasks without QQP, as they may affect the semantic similarity. For human-crafted examples, we apply \textit{CheckList} (C8) to SST-2, QQP, and QNLI; \textit{StressTest} (C9) and \textit{ANLI} (C10) to MNLI; and \textit{AdvSQuAD} (C11) to QNLI tasks.}
    \label{tab:stats}
\setlength{\tabcolsep}{3.5pt}
    \begin{tabular}{lcccccccccccccc}
    \toprule
      \multirow{2}{*}{\textbf{Corpus}}  & \multirow{2}{*}{\textbf{Task}} & \textbf{|Train|} & {\textbf{|Test|}} & \multicolumn{5}{c}{\textbf{Word-Level }} & \multicolumn{2}{c}{\textbf{Sent.-Level}} & \multicolumn{4}{c}{\textbf{Human-Crafted }}   \\   
      & & \scriptsize{(GLUE)} &  \scriptsize{(AdvGLUE)}  &  C1 & C2 & C3 & C4 & C5 & C6 & C7 & C8 & C9 & C10 & C11  \\
        \midrule
        \textbf{SST-2} & sentiment & 67,349 & 1,420 & 204 & 197 & 91 & 175 & 64 & 211 & 320 & 158 & 0 & 0 & 0  \\
        \textbf{QQP}  & paraphrase & 363,846 & 422 & 42 & 151 & 17 & 35 & 75 & 37 & 0 & 65 & 0 & 0 & 0 \\
        \textbf{QNLI} & NLI/QA & 104,743 & 968 & 73 & 139 & 71 & 98 & 72 & 159 & 219 & 80 & 0 & 0 & 57 \\
        \textbf{RTE}  & NLI & 2,490 & 304 & 43 & 44 & 31 & 27 & 23 & 48 & 88 & 0 & 0 & 0 & 0 \\
        \textbf{MNLI} & NLI & 392,702 & 1,864 & 69 & 402 & 114 & 161 & 128 & 217 & 386 & 0 & 194 & 193 & 0 \\
        \midrule
        \multicolumn{3}{l}{\textbf{Sum of \method test set}} & 4,978 & 431 & 933 & 324 & 496 & 362 & 672 & 1013 & 303 & 194 & 193 & 57 \\
        \bottomrule
        \end{tabular}
\end{table}

\section{Dataset Construction}

In this section, we provide an overview of our evaluation tasks, as well as the pipeline of how we construct the benchmark data.
During this data construction process, we also compare the effectiveness of different adversarial attack methods, and present several interesting findings.

\subsection{Overview}

\textbf{Tasks.} We consider the following five most representative and challenging tasks used in GLUE~\citep{wang2018glue}:  Sentiment Analysis (\textit{SST-2}),  
Duplicate Question Detection (\textit{QQP}), and 
Natural Language Inference (NLI, including \textit{MNLI, RTE, QNLI}).
The detailed explanation for each task can be found in Appendix \ref{appendix:tasks}. Some tasks in GLUE are not included in  \method, since
there are either no well-defined automatic adversarial attacks (\emph{e.g.}, \textit{CoLA}), or insufficient data  (\emph{e.g.}, \textit{WNLI}) for the attacks.

\textbf{Dataset Statistics and Evaluation Metrics.} 
\method follows the same training data and evaluation metrics as GLUE. In this way, models trained on the GLUE training data can be easily evaluated under IID sampled test sets (GLUE benchmark) or carefully crafted adversarial test sets (\method benchmark). 
Practitioners can understand the model generalization via the GLUE diagnostic test suite and examine the model robustness against different levels of adversarial attacks from the \method diagnostic report  with only one-time training.
Given the same evaluation metrics, model developers can clearly understand the performance gap between models tested in the ideally benign environments and  approximately worst-case adversarial scenarios.
We present the detailed dataset statistics under various attacks in Table \ref{tab:stats}. Detailed label distribution and evaluation metrics are  in Appendix Table \ref{tab:labeldist}.

\subsection{Adversarial Perturbations}
In this section, we detail how we optimize different levels of  adversarial perturbations to the benign source samples and collect the raw adversarial data with noisy labels, which will then be carefully filtered by human annotators described in the next section.
Specifically, we consider the dev sets of GLUE benchmark as our source samples, upon which we perform different adversarial attacks. For relatively large-scale tasks (QQP, QNLI, MNLI-m/mm), we sample 1,000 cases from the dev sets for efficiency purpose.
For the remaining tasks, we consider the whole dev sets as source samples.

\subsubsection{Word-level Perturbation}
Existing word-level adversarial attacks perturb the words through different strategies, such as perturbing words with their synonyms \citep{textfooler} or carefully crafted typo words \citep{textbugger} (\emph{e.g.}, ``foolish'' to ``fo01ish''), such that the perturbation does not change the semantic meaning of the sentences but dramatically change the models' output. 
To examine the model robustness against different perturbation strategies, we select one representative adversarial attack method for each strategy as follows.

\textbf{Typo-based Perturbation.}  We select TextBugger \citep{textbugger} as the representative algorithm for generating typo-based adversarial examples. 
When performing the attack, TextBugger first identifies the important words and then replaces them with typos.

\textbf{Embedding-similarity-based Perturbation.} 
We choose TextFooler \citep{textfooler} as the representative adversarial attack that considers embedding similarity as a constraint to generate semantically consistent adversarial examples. Essentially, TextFooler first performs word importance ranking, and then substitutes those important ones to their synonyms extracted according to the cosine similarity of word embeddings.

\begin{figure}
    \centering
    \includegraphics[width=\linewidth]{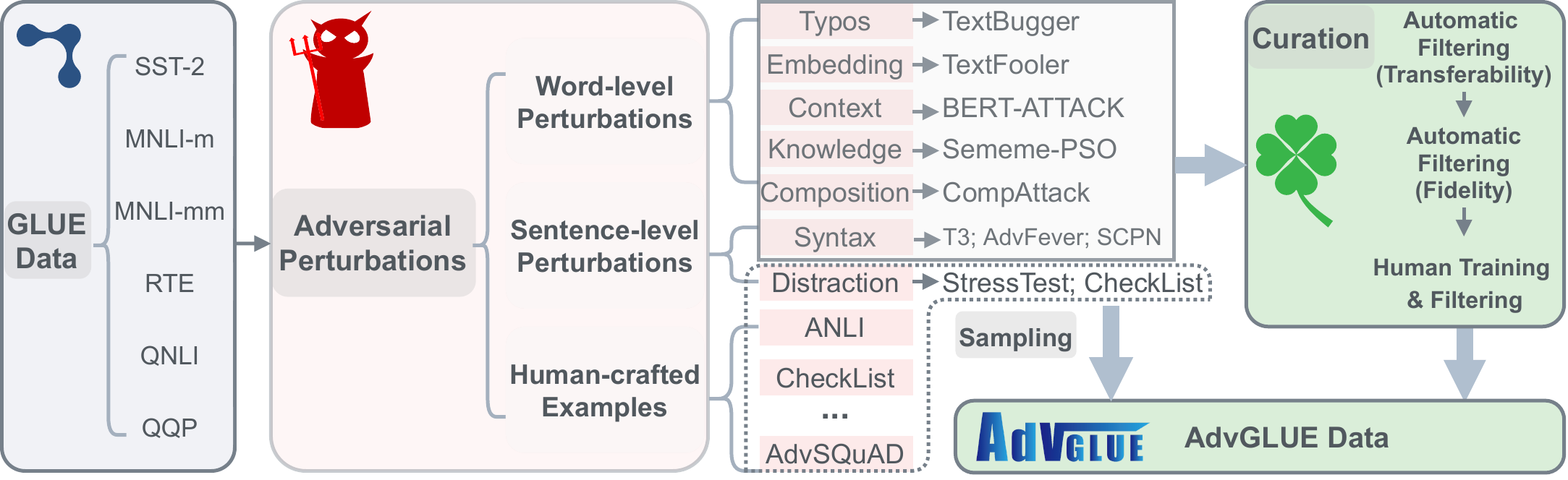}
    \caption{\small Overview of the AdvGLUE dataset construction pipeline.}
    \label{fig:overview}
\end{figure}

\textbf{Context-aware Perturbation.} We use BERT-ATTACK \citep{bertattack} to generate context-aware perturbations. 
The fundamental difference between BERT-ATTACK and TextFooler lies on the word replacement procedure. Specifically, BERT-ATTACK uses the pre-trained BERT to perform masked language prediction to generate contextualized potential word replacements for those crucial words.

\textbf{Knowledge-guided Perturbation.} We consider SememePSO \citep{comattack} as an example to generate adversarial examples guided by the HowNet \citep{hownet} knowledge base. 
SememePSO first finds out substitutions for each word in HowNet based on sememes, and then searches for the optimal combination based on particle swarm optimization. 

\textbf{Compositions of different Perturbations.} We also implement a whitebox-based adversarial attack algorithm called CompAttack that integrates the aforementioned perturbations in one algorithm to evaluate model robustness to various adversarial transformations. Moreover, we efficiently search for perturbations via optimization so that CompAttack can achieve the attack goal while perturbing the minimal number of words. The implementation details can be found in Appendix \ref{appendix:compattack}.

We note that the above adversarial attacks require a surrogate model to search for the optimal perturbations. In our experiments, we follow the setup of ANLI \citep{anli} and generate adversarial examples against three different types of  models (BERT, RoBERTa, and RoBERTa ensemble) trained on the GLUE benchmark. 
We then perform one round of filtering to retain those examples with high \textit{adversarial transferability} between these surrogate models. 
We discuss more implementation details and hyper-parameters of each attack method in Appendix \ref{appendix:imple}.

\subsubsection{Sentence-level Perturbation}
Different from word-level attacks that perturb specific words, sentence-level attacks mainly focus on the syntactic and logical structures of  sentences. Most of them achieve the attack goal by either paraphrasing the sentence,  manipulating the syntactic structures, or inserting some unrelated sentences to distract the model
attention. \method considers the following representative perturbations.

\textbf{Syntactic-based Perturbation.} We incorporate three adversarial attack strategies that manipulate the sentence based on the syntactic structures. ($i$) \textit{Syntax Tree Transformations}. SCPN \citep{scpn} is trained to produce a paraphrase of a given sentence with specified syntactic structures. Following the default setting, we select the most frequent $10$ templates from ParaNMT-50M corpus \cite{wieting2017paranmt} to guide the generation process. An LSTM-based encoder-decoder model (SCPN) is used to generate parses of target sentences according to the templates. These parses are further fed into another SCPN to generate full sentences. We use the pre-trained SCPNs released by the official codebase. 
($ii$) \textit{Context Vector Transformations}. T3 \citep{t3} is a whitebox attack algorithm that can add perturbations on different levels of the syntax tree and generate the adversarial sentence.
In our setting, we add perturbations to the context vector of the root node given syntax tree, which is iteratively optimized to construct the adversarial sentence. 
($iii$) \textit{Entailment Preserving Transformations}. We follow the entailment preserving rules proposed by AdvFever \citep{advfever}, and transform all the sentences satisfying the templates into semantically equivalent ones. More details can be found in Appendix \ref{appendix:imple}. 

\textbf{Distraction-based Perturbation.} We integrate two attack strategies: ($i$) StressTest \citep{stresstest} appends three true statements (``and true is true'', ``and false is not true'', ``and true is true'' for five times) to the end of the hypothesis sentence for NLI tasks. 
($ii$) CheckList \citep{checklist}  adds randomly generated URLs and handles to distract model attention. 
Since the aforementioned distraction-based perturbations may impact the linguistic acceptability and the understanding of semantic equivalence, we mainly apply these rules to part of the GLUE tasks, including \textit{SST-2} and NLI tasks (\textit{MNLI, RTE, QNLI}), to evaluate whether model can be easily misled by the strong negation words or such lexical similarity.

\begin{table}[t]\small \setlength{\tabcolsep}{7pt}
\centering
\caption{\small \textbf{Examples of \method benchmark}. We show $3$ examples from QNLI task. These examples are generated with three levels of perturbations and they all can successfully change the predictions of all surrogate models (BERT, RoBERTa and RoBERTa ensemble).}

 \label{tab:examples}
\resizebox{1.0\textwidth}{!}{
\begin{tabular}{p{2.0cm}p{9cm}p{1.5cm}}
\toprule 
Linguistic \quad Phenomenon & Samples (\st{Strikethrough} = Original Text, \textcolor{red}{red} = Adversarial Perturbation) & Label $\rightarrow$ Prediction \\
\midrule
 \multirow{4}{*}{\shortstack{Typo \\ (Word-level)}} & \textbf{Question}: What was the population of the Dutch Republic before this emigration? & \multirow{4}{*}{False $\rightarrow$ True} \\
 & \textbf{Sentence}: This was a \st{huge} \textcolor{red}{hu ge} influx as the entire population of the Dutch Republic amounted to ca. & \\
  \midrule
 \multirow{4}{*}{\shortstack{Distraction \\ (Sent.-level)}} & \textbf{Question}: What was the population of the Dutch Republic before this emigration? \textcolor{red}{https://t.co/DlI9kw} & \multirow{4}{*}{False $\rightarrow$ True} \\
 & \textbf{Sentence}: This was a huge influx as the entire population of the Dutch Republic amounted to ca. & \\
  \midrule
 \multirow{3}{*}{\shortstack{CheckList\\ (Human-crafted)}} & \textbf{Question}: What is Tony's profession? & \multirow{3}{*}{True  $\rightarrow$ False} \\
 & \textbf{Sentence}: Both Tony and Marilyn were executives, but there was a change in Marilyn, who is now an assistant. & \\
\bottomrule
\end{tabular}
}
\end{table}

\subsubsection{Human-crafted Examples}
To ensure  our benchmark covers more linguistic phenomena in addition to those provided by automatic attack algorithms, we integrate the following high-quality human-crafted adversarial data from crowd-sourcing or expert-annotated templates and transform them to the formats of GLUE tasks. 

\textbf{CheckList}\footnote{We note that both CheckList and StressTest propose both rule-based distraction sentences and manually crafted templates to generate test samples. The former is considered as sentence-level distraction-based perturbations, while the latter is considered as human-crafted examples. } \citep{checklist} is a testing method designed for analysing different capabilities of NLP models using different test types. For each task, CheckList first identifies necessary natural language capabilities a model should have, then designs several test templates to generate test cases at scale. We follow the instructions and collect testing cases for three tasks: \textit{SST-2, QQP} and \textit{QNLI}. For each task, we adopt two capability tests: \textit{Temporal} and \textit{Negation}, which test if the model understands the order of events and if the model is sensitive to negations. %

\textbf{StressTest}\footnotemark[\value{footnote}] \citep{stresstest} proposes carefully crafted rules to construct ``stress tests'' and evaluate robustness of NLI models to specific linguistic phenomena. We adopt the test cases focusing on \textit{Numerical Reasoning} into our adversarial \textit{MNLI} dataset. These premise-hypothesis pairs are able to test whether the model can perform reasoning involving numbers and quantifiers and predict the correct relation between premise and hypothesis. 

\textbf{ANLI} \citep{anli} is a large-scale NLI dataset collected iteratively in a human-in-the-loop manner. In each iteration, human annotators are asked to design sentences to fool current model. Then the model is further finetuned on a larger dataset incorporating these sentences, which leads to a stronger model. Finally, annotators are asked to write harder examples to detect the weakness of this stronger model. In the end, the sentence pairs generated in each round form a comprehensive dataset that aims at examining the vulnerability of NLI models. We adopt ANLI into our adversarial \textit{MNLI} dataset. We obtain the permission from the ANLI authors to include the ANLI dataset as part of our leaderboard.

\textbf{AdvSQuAD} \citep{advsquad} is an adversarial dataset targeting at reading comprehension systems. Adversarial examples are generated by appending a distracting sentence to the end of the input paragraph. The distracting sentences are carefully designed to have common words with questions and look like a correct answer to the question.
We mainly consider the examples generated by \textsc{AddSent} and \textsc{AddOneSent} strategies, and adopt the distracting sentences and questions in the \textit{QNLI} format with labels ``not answered''.
The use of AdvSQuAD in \method is authorized by the authors.

We present sampled \method examples with the word-level, sentence-level perturbations and human-crafted samples in Table \ref{tab:examples}. More examples are provided in  Appendix \ref{appendix:examples}.

\subsection{Data Curation}
After collecting the raw adversarial dataset, additional rounds of filtering are required to guarantee its quality and validity. We consider two types of filtering: automatic filtering and human evaluation. 

\textbf{Automatic Filtering} mainly evaluates the generated adversarial examples along two fronts: \textit{transferability} and \textit{fidelity}.
\begin{enumerate}[leftmargin=*]
\item \textbf{Transferability} evaluates whether the adversarial examples generated against one source model (\emph{e.g.}, BERT) can successfully transfer and attack the other two (\emph{e.g.}, RoBERTa and RoBERTa ensemble), given the surrogate models used to generate adversarial examples (BERT, RoBERTa and RoBERTa ensemble).
Only  adversarial examples that can successfully transfer to the other two models will be kept for the next round of fidelity filtering, so that the selected examples can exploit the biases shared across different models and unveil their fundamental  weakness.
\item \textbf{Fidelity} evaluates how the generated adversarial examples maintain the original semantics. For word-level adversarial examples, we use \textit{word modification rate} to measure what percentage of words are perturbed. Concretely, word-level adversarial examples with word modification rate larger than $15\%$ are filtered out. 
For sentence-level adversarial examples, we use \textit{BERTScore} \citep{bertscore} to evaluate the semantic similarity between the adversarial sentences and their corresponding original ones. For each sentence-level attack, adversarial examples with the highest similarity scores are kept to guarantee their semantic closeness to the benign samples.
\end{enumerate}

\begin{table}[t]
\small
\centering
    \caption{\small \textbf{Statistics of data curation}. We report  Attack Success Rate (\textbf{ASR}) and  ASR after data curation (\textbf{Curated ASR}) to evaluate the \textit{effectiveness} of different adversarial attacks. We  present the \textbf{Filter Rate} of data curation and inter-annotator agreement rate (\textbf{Fleiss Kappa}) before and after curation to evaluate the \textit{validity} of adversarial examples. \textbf{Human Accuracy} on our curated dataset is evaluated by taking one random annotator's annotation as prediction and the majority voted label as ground truth. SPSO: SememePSO, TF: TextFooler, TB:TextBugger, CA: CompAttack, BA:BERT-ATTACK. $\uparrow$/$\downarrow$: higher/lower the better.}
    \label{tab:curation}
{
{
\setlength{\tabcolsep}{3.75pt}
    \begin{tabular}{ll|cccccccc|c}
    \toprule
         \multirow{2}{*}{\textbf{Tasks}} & \multirow{2}{*}{\textbf{Metrics}} & \multicolumn{5}{c}{Word-level Attacks} & \multicolumn{3}{c|}{Sentence-level Attacks} & \multirow{2}{*}{\textbf{Avg}} \\
         \cmidrule(lr){3-7} \cmidrule(lr){8-10}
         & & SPSO & TF & TB & CA &  BA & T3 & SCPN & AdvFever &    \\
        \midrule
          \multirow{6}{*}{SST-2} & ASR $\uparrow$ & 89.08 & 95.38 & 88.08 & 31.91 & 39.77 & \textbf{97.69} & 65.37 & 0.57 & 63.48 \\
           & Curated ASR $\uparrow$ & 8.29 & 8.97 & 8.85 & 4.02 & 4.04 & \textbf{10.45} & 6.88 & 0.23 & 6.47 \\
           & Filter Rate $\downarrow$ & 90.71 & 90.62 & 90.04 & 86.63 & 89.81 & 89.27 & 89.47 & \textbf{60.00} & 85.82 \\
           & Fleiss Kappa $\uparrow$  & 0.22 & 0.20 & \textbf{0.50} & 0.21 & 0.24 & 0.23 & 0.29 & 0.12 & 0.26 \\
           & Curated Fleiss Kappa $\uparrow$  & 0.51 & 0.49 & \textbf{0.67} & 0.46 & 0.45 & 0.44 & 0.47 & 0.20 & 0.52 \\
           & Human Accuracy $\uparrow$ & 0.85 & 0.86 & \textbf{0.91} & 0.88 & 0.85 & 0.78 & 0.85 & 0.50 & 0.87 \\
           \midrule
           \multirow{6}{*}{MNLI} & ASR $\uparrow$ & 78.45 & 61.50 & 69.35 & 68.58 & 65.02 & \textbf{91.23} & 87.73 & 2.25 & 65.51 \\
           & Curated ASR $\uparrow$ & 3.48 & 1.55 & \textbf{8.94} & 3.11 & 2.58 & 3.41 & 6.75 & 0.30 & 3.77 \\
           & Filter Rate $\downarrow$ & 95.59 & 97.55 & 87.12 & 95.45 & 96.10 & 96.27 & 92.31 & \textbf{86.63} & 93.38 \\
           & Fleiss Kappa $\uparrow$  & 0.28 & 0.24 & \textbf{0.53} & 0.39 & 0.32 & 0.28 & 0.24 & 0.35 & 0.33 \\
           & Curated Fleiss Kappa $\uparrow$  & 0.65 & 0.59 & \textbf{0.74} & 0.65 & 0.60 & 0.56 & 0.60 & 0.51 & 0.67 \\
           & Human Accuracy $\uparrow$ & 0.85 & 0.83 & \textbf{0.91} & 0.89 & 0.83 & 0.84 & \textbf{0.91} & 0.83 & 0.89 \\
           \midrule
           \multirow{6}{*}{RTE} & ASR $\uparrow$ & 76.67 & 75.67 & 85.89 & 73.36 & 72.05 & \textbf{92.39} & 88.45 & 6.62 & 71.39 \\
           & Curated ASR $\uparrow$ & 6.20 & 8.14 & \textbf{10.03} & 6.97 & 5.58 & 7.05 & 8.30 & 2.53 & 6.85 \\
           & Filter Rate $\downarrow$ & 91.93 & 89.21 & 88.29 & 90.72 & 92.16 & 92.31 & 90.61 & \textbf{61.34} & 87.07 \\
           & Fleiss Kappa $\uparrow$  & 0.30 & 0.32 & \textbf{0.58} & 0.35 & 0.25 & 0.33 & 0.43 & \textbf{0.58} & 0.38 \\
           & Curated Fleiss Kappa $\uparrow$  & 0.49 & 0.67 & \textbf{0.80} & 0.63 & 0.42 & 0.60 & 0.64 & 0.65 & 0.66 \\
           & Human Accuracy $\uparrow$ & 0.77 & \textbf{0.95} & 0.94 & 0.87 & 0.79 & 0.89 & 0.91 & 0.86 & 0.92 \\
           \midrule
           \multirow{6}{*}{QNLI} & ASR $\uparrow$ & 71.88 & 67.03 & 82.54 & 67.24 & 60.53 & \textbf{96.41} & 67.37 & {0.97} & 64.25 \\
           & Curated ASR  $\uparrow$& 3.92 & 2.87 & 5.87 & 4.09 & 2.69 & \textbf{7.59} & 3.90 & 0.00 & 3.87 \\
           & Filter Rate $\downarrow$ & 94.63 & 95.89 & 92.89 & 93.92 & 95.78 & \textbf{92.16} & 94.21 & 100.00 & 94.93 \\
           & Fleiss Kappa $\uparrow$  & 0.07 & 0.05 & \textbf{0.16} & 0.10 & 0.14 & 0.07 & 0.12 & -0.16 & 0.11 \\
           & Curated Fleiss Kappa $\uparrow$  & 0.37 & 0.43 & 0.49 & 0.34 & \textbf{0.53} & 0.37 & 0.43 & - & 0.44 \\
           & Human Accuracy $\uparrow$ & 0.80 & 0.86 & 0.85 & 0.82 & \textbf{0.92} & 0.89 & \textbf{0.92} & - & 0.85 \\
           \midrule
           \multirow{6}{*}{QQP} & ASR $\uparrow$ & 45.86 & 48.59 & \textbf{57.92} & 49.33 & 43.66 & 48.20 & 44.37 & 0.30 & 42.28 \\
           & Curated ASR $\uparrow$ & 1.52 & 1.74 & \textbf{5.87} & 3.05 & 0.76 & 1.47 & 1.50 & 0.00 & 1.99 \\
           & Filter Rate $\downarrow$ & 96.73 & 96.50 & \textbf{89.90} & 93.83 & 98.28 & 97.04 & 96.62 & 100.00 & 96.11 \\
           & Fleiss Kappa $\uparrow$  & 0.26 & 0.27 & \textbf{0.38} & 0.27 & 0.24 & 0.25 & 0.29 & - & 0.30 \\
           & Curated Fleiss Kappa $\uparrow$  & 0.32 & 0.46 & \textbf{0.62} & 0.48 & 0.40 & 0.10 & 0.47 & - & 0.51 \\
           & Human Accuracy $\uparrow$ & 0.84 & 0.98 & 0.97 & 0.89 & 0.78 & 0.89 & \textbf{1.00} & - & 0.89 \\
           \bottomrule
    \end{tabular}
    \vspace{-3mm}
  }
}
\end{table}

\textbf{Human Evaluation} validates whether the adversarial examples preserve the original labels and whether the labels are highly agreed among annotators. Concretely, we recruit annotators from Amazon Mechanical Turk. To make sure the annotators fully understand the GLUE tasks, each worker is required to pass  a training step to be qualified to work on the main filtering tasks for the generated adversarial examples. We tune the pay rate for different tasks, as shown in Appendix Table \ref{tab:humanstat}. The pay rate of the main filtering phase is twice as much as that of the training phase.
\begin{enumerate}[leftmargin=*]
\item \textbf{Human Training Phase} is designed to ensure that the annotators understand the tasks. The annotation instructions for each task follows \citep{muppet}, and we provide at least two examples for each class to help annotators understand the tasks.\footnote{Instructions can be found at \url{https://adversarialglue.github.io/instructions}.} Each annotator is required to work on a batch of 20 examples randomly sampled from the GLUE dev set. After  annotators answer each example, a ground-truth answer will be provided to help them understand whether the answer is correct. Workers who get at least $85\%$ of the examples correct during training are qualified to work on the main filtering task. A total of 100 crowd workers participated in each task, and the number of qualified workers are shown in Appendix Table \ref{tab:humanstat}. We also test the human accuracy of qualified annotators for each task on 100 randomly sampled examples from the dev set excluding the training samples. The details and results can be found in Appendix Table \ref{tab:humanstat}.
\item \textbf{Human Filtering Phase} verifies the quality of the generated adversarial examples and only maintains high-quality ones to construct the benchmark dataset. Specifically, annotators are required to work on a batch of 10 adversarial examples generated from the same attack method. Every adversarial example will be validated by 5 different annotators.
Examples are selected following two criteria: ($i$) high consensus: each example must have at least 4-vote consensus; ($ii$) utility preserving: the majority-voted label must be the same as the original one to make sure the attacks are valid (\emph{i.e.}, cannot fool human) and preserve the semantic content. 
\end{enumerate}

The data curation results including inter-annotator agreement rate (Fleiss Kappa) and human accuracy on the curated dataset are shown in Table \ref{tab:curation}. 
We will provide more analysis in the next section.
Note that even after the data curation step, some grammatical errors and typos can still remain, as some adversarial attacks intentionally inject typos (\emph{e.g.}, TextBugger) or manipulate syntactic trees (\emph{e.g.}, SCPN) which are very stealthy. We will retain these samples as their labels receive high consensus from annotators, which means the typos do not substantially impact humans' understanding.

\begin{table}[t]
\small
    \centering
    \caption{\small \textbf{Model performance on \method test set}. BERT (Large) and RoBERTa (Large) are fine-tuned using different random seeds and thus different from the surrogate models used for adversarial text generation. 
    For MNLI, we report the test accuracy on the matched  and mismatched test sets; for QQP, we report accuracy and F1; and for other tasks, we report the accuracy. All values are reported by percentage (\%). We also report the macro-average (Avg) of per-task scores for different models. (Complete results are listed in our leaderboard.)}
    \label{tab:benchmark}
\setlength{\tabcolsep}{3.75pt}
    \begin{tabular}{l|ccccc|ccc}
    \toprule
    \multirow{2}{*}{\textbf{Model}} & \multirow{1}{*}{\textbf{SST-2}} &  \multirow{1}{*}{\textbf{MNLI}} &  \multirow{1}{*}{\textbf{RTE}} &  \multirow{1}{*}{\textbf{QNLI}} &   \multirow{1}{*}{\textbf{QQP}} & \textbf{Avg} & \textbf{Avg} & \textbf{Avg}   \\
         & \scriptsize{AdvGLUE} & \scriptsize{AdvGLUE} & \scriptsize{AdvGLUE} & \scriptsize{AdvGLUE} & \scriptsize{AdvGLUE} & \scriptsize{AdvGLUE} & \scriptsize{GLUE} & \scriptsize{$\Delta$} $\downarrow$ \\
    \midrule 
    \multicolumn{8}{l}{\emph{State-of-the-art Pre-trained Language Models}} \\
    \midrule
    {BERT (Large)} & 33.03 & 28.72/27.05 & 40.46 & 39.77 & 37.91/16.56 & 33.68 & 85.76  & 52.08 \\
    {ELECTRA (Large)} & 58.59 & 14.62/20.22 & 23.03 & 57.54 & 61.37/42.40  & 41.69 & \textbf{93.16} & 51.47 \\
    {RoBERTa (Large)} & 58.52 & 50.78/39.62 & 45.39 & 52.48 & 57.11/41.80  & 50.21 & 91.44 & 41.23 \\
    \m{{T5 (Large)}} & \m{60.56} & \m{48.43/38.98} & \m{62.83} & \m{57.64} & \m{63.03/\textbf{55.68}} & \m{56.82} & \m{90.39} & \m{33.57} \\
    {ALBERT (XXLarge)} & \textbf{66.83} & 51.83/44.17 & 73.03 & \textbf{63.84} & 56.40/32.35  & 59.22 & 91.87 & 32.65 \\
    \m{{DeBERTa (Large)}} & \m{57.89} & \m{\textbf{58.36/52.46}} & \m{\textbf{78.95}} & \m{57.85} & \m{60.43/47.98} & \m{\textbf{60.86}} & \m{92.67} & \m{\textbf{31.81}} \\
    \midrule 
    \multicolumn{8}{l}{\emph{Robust Training Methods for Pre-trained Language Models}} \\
    \midrule 
    {SMART (BERT)} & 25.21 & 26.89/23.32 & 38.16 & 34.61 & 36.49/20.24  & 30.29 & 85.70 & 55.41\\
    {SMART (RoBERTa)} & 50.92 & 45.56/36.07 & 70.39 & 52.17 & \textbf{64.22}/44.28  & 53.71 & 92.62 & 38.91 \\
    {FreeLB (RoBERTa)} & 61.69 & 31.59/27.60 & 62.17 & 62.29 & 42.18/31.07  & 50.47 & 92.28 & 41.81 \\
    {InfoBERT (RoBERTa)} & 47.61 & 50.39/41.26 & 39.47 & 54.86 & 49.29/35.54  & 46.04 & 89.06 & 43.02 \\
    \bottomrule
    \end{tabular}
\end{table}

\subsection{Benchmark of Adversarial Attack Algorithms}
\label{sec:attack}
Our data curation phase also \m{serves} as a comprehensive benchmark over existing  adversarial attack methods, as it provides a fair standard for all adversarial attacks and systematic human annotations to evaluate the quality of the generated samples.

\textbf{Evaluation Metrics.} Specifically, we evaluate these attacks along two fronts: \textit{effectiveness} and \textit{validity}.
For effectiveness, we consider two evaluation metrics: \textbf{Attack Success Rate (ASR)} and \textbf{Curated Attack Success Rate (Curated ASR)}.  
Formally, given a benign dataset $\mathcal{D}=\{(x^{(i)}, y^{(i)})\}^N_{i=1}$ consisting of $N$ pairs of sample $x^{(i)}$ and ground truth $y^{(i)}$, for an adversarial attack method $\mathcal{A}$ that generates an adversarial example $\mathcal{A}(x)$ given an input $x$ to attack a surrogate model $f$, ASR is calculated as
\begin{equation}\small
    \textbf{ASR} =  \sum_{(x,y) \in \mathcal{D}} \frac{\mathbb{1}[f(\mathcal{A}(x)) \neq y]}{\mathbb{1}[f(x) = y]},
\end{equation}
where $\mathbb{1}$ is the indicator function.
After the data curation phase, we collect a curated adversarial dataset $\mathcal{D}_c$. Thus, Curated ASR is calculated as 
\begin{equation}\small
    \textbf{Curated ASR} =  \sum_{(x,y) \in \mathcal{D}} \frac{\mathbb{1}[f(\mathcal{A}(x)) \neq y] \cdot \mathbb{1}[\mathcal{A}(x) \in \mathcal{D}_c]}{\mathbb{1}[f(x) = y]}.
\end{equation}

For validity, we consider three evaluation metrics: \textbf{Filter Rate}, \textbf{Fleiss Kappa}, and \textbf{Human Accuracy}. Specifically, Filter Rate is calculated by $1 - \frac{\textrm{Curated ASR}}{\textrm{ASR}}$ to measure how many examples are rejected in the data curation procedures and can reflect the noisiness of the generated adversarial examples. 
We report the average ASR, Curated ASR, and Filter Rate over the three surrogate models we consider in Table \ref{tab:curation}.
\m{Fleiss Kappa is a widely used metric in existing datasets (\emph{e.g.}, SNLI, ANLI, and FEVER \citep{snli, anli, fever}) to measure the inter-annotator agreement rate on the collected dataset. Fleiss Kappa between 0.4 and 0.6 is considered as moderate agreement and between 0.6 and 0.8 as substantial agreement. The inter-annotator agreement rates of most high-quality datasets fall into these two intervals. In this paper, we follow the standard protocol and report Fleiss Kappa and Curated Fleiss Kappa to analyze the inter-annotator agreement rate on the collected adversarial dataset before and after curation to reflect the ambiguity of  generated  examples. We also estimate the human performance on our curated datasets. Specifically, given a sample with 5 annotations, we take one random annotator's annotation as the prediction and the majority voted label as the ground truth and calculate the human accuracy as shown in Table \ref{tab:curation}.} 

\textbf{Analysis.} As shown in Table \ref{tab:curation}, in terms of attack \textit{effectiveness}, while most attacks show high ASR, the Curated ASR is always less than $11\%$, which indicates that most existing adversarial attack algorithms are not effective enough to generate high-quality adversarial examples. 
In terms of \textit{validity}, the filter rates for most adversarial attack methods are more than $85\%$, which suggests that existing strong adversarial attacks are prone to generating invalid adversarial examples that either change the original semantic meanings or generate ambiguous perturbations that hinder the annotators' unanimity. We provide detailed filter rates for automatic filtering and human evaluation in Appendix Table \ref{tab:filter}, and the conclusion is that around $60-80\%$ of examples are filtered due to the low transferability and high word modification rate. Among the remaining samples, around $30-40\%$ examples are filtered due to the low human agreement rates (Human Consensus Filtering), and around $20-30\%$ are filtered due to the semantic changes which lead to the label changes (Utility Preserving Filtering). We also note that the data curation procedures are indispensable for the adversarial evaluation, as the Fleiss Kappa before curation is very low, suggesting that a lot of adversarial sentences have unreliable labels and thus tend to underestimate the model robustness against the textual adversarial attacks. After the data curation, our \method shows a Curated Fleiss Kappa of near 0.6,
comparable with existing high-quality dataset such as SNLI and ANLI.  
Among all the existing attack methods, we observe that TextBugger is the most effective and valid attack method, as it demonstrates the highest Curated ASR and Curated Fleiss Kappa across different  tasks.

\subsection{Finalizing the Dataset}
The full pipeline of constructing  \method is summarized in Figure \ref{fig:overview}.

\textbf{Merging.}  We note that distraction-based adversarial examples and human-crafted adversarial examples are guaranteed to be valid by definition or crowd-sourcing annotations, and thus data curation is not needed on these attacks. When merging them with our curated set, 
we calculate the average number of samples per attack from our curated set, and sample the same amount of adversarial examples from these attacks following the same label distribution. This way, each  attack contributes to similar amount of adversarial data, so that \method can evaluate models against different types of  attacks with similar weights and provide a comprehensive and unbiased diagnostic report.

\textbf{Dev-Test Split.} After collecting the adversarial examples from the considered  attacks, we split the final dataset into a dev set and a test set. In particular, we first randomly split the benign data into $9:1$, and the adversarial examples generated based on $90\%$ of the benign data serve as the hidden test set, while the others are published as the dev set. For human-crafted adversarial examples, since they are not generated based on the benign GLUE data, we randomly select $90\%$ of the data as the test set, and the remaining $10\%$ as the dev set.
The dev set is publicly released to help participants to understand the tasks and the data format. To protect the integrity of our test data, the test set will not be released to the public. Instead, participants are required to upload the model to CodaLab, which automates the evaluation process on the hidden test set and provides a diagnostic report.

\begin{table}[t]
\small
    \centering
    \caption{\small \textbf{Diagnostic report of state-of-the-art language models and robust training methods}. For each attack method, we evaluate models against generated adversarial data for different tasks to obtain per-task accuracy scores, and report the macro-average of those scores. (C1=\textit{Embedding-similarity}, C2=\textit{Typos}, C3=\textit{Context-aware}, C4=\textit{Knowledge-guided}, C5=\textit{Compositions}, C6=\textit{Syntactic-based Perturbations},
    C7=\textit{Distraction-based Perturbations},
    C8=\textit{CheckList},
    C9=\textit{StressTest},
    C10=\textit{ANLI} and 
    C11=\textit{AdvSQuAD}).}
    \label{tab:report}
\setlength{\tabcolsep}{3.75pt}
    \begin{tabular}{l|ccccc|cc|cccc}
    \toprule
    \multirow{2}{*}{\textbf{Models}} & \multicolumn{5}{c|}{\textbf{Word-Level Perturbations}} & \multicolumn{2}{c|}{\textbf{Sent.-Level}} &  \multicolumn{4}{c}{\textbf{Human-Crafted Examples}}   \\
     & C1 & C2 & C3 & C4 & C5 & C6 & C7 & C8 & C9 & C10 & C11 \\
    \midrule 
    {BERT (Large)} & 42.02 & 31.96 & 45.18 & 45.86 & 33.85 & 44.86 & 24.16 & 16.33 & 23.20 & 13.47 & 10.53 \\
   {ELECTRA (Large)} & 43.07 & 45.12 & 47.95 & 46.33 & 47.33 & 43.47 & 33.30 & 32.20 & 26.29 & 26.94 & 52.63 \\
    {RoBERTa (Large)} & 56.54 & 57.19 & 60.47 & 49.81 & 55.92 & 50.49 & 41.89 & 37.78 & 28.35 & 16.58 & 35.09 \\
    \m{{T5 (Large)}} & \m{60.04} & \m{67.94} & \m{64.60} & \m{59.84} & \m{58.50} & \m{50.54} & \m{42.20} & \m{\textbf{69.02}} & \m{23.20} & \m{17.10} & \m{52.63} \\
    {ALBERT (XXLarge)} & \textbf{66.71} & 67.61 & \textbf{73.49} & \textbf{70.36} & 59.52 & \textbf{63.76} & \textbf{49.14} & 45.55 & 39.69 & 26.94 & 43.86 \\
    \m{{DeBERTa (Large)}} & \m{65.07} & \m{\textbf{74.87}} & \m{68.02} & \m{65.30} & \m{\textbf{62.54}} & \m{57.41} & \m{47.22} & \m{45.08} & \m{\textbf{52.06}} & \m{22.80} & \m{54.39} \\
   \midrule
    {SMART (BERT)} & 45.17 & 31.04 & 42.89 & 45.23 & 30.76 & 40.74 & 16.62 & 8.20 & 18.56 & 10.36 & 1.75 \\
    {SMART (RoBERTa)} & 62.93 & 58.03 & 65.09 & 62.65 & 61.37 & 55.31 & 40.13 & 39.27 & 28.35 & 15.54 & 31.58 \\
    {FreeLB (RoBERTa)} & 51.95 & 53.23 & 52.92 & 51.15 & 52.18 & 50.75 & 37.72 & 66.87 & 23.71 & \textbf{29.02} &\textbf{ 64.91} \\
    {InfoBERT (RoBERTa)} & 55.47 & 55.78 & 59.02 & 51.33 & 55.48 & 44.56 & 31.49 & 34.31 & 42.27 & 14.51 & 43.86 \\
    \bottomrule
    \end{tabular}
\end{table}

\section{Diagnostic Report for Language Models}

\textbf{Benchmark Results.}
We follow the official implementations and training scripts of \m{pre-trained} language models to reproduce results on GLUE and test these models on AdvGLUE. The training details can be found in Appendix \ref{appendix:train}.
Results are summarized in Table \ref{tab:benchmark}. We observe that although state-of-the-art language models have achieved high performance on GLUE, they are vulnerable to various adversarial attacks. For instance, the performance gap can be as large as $55\%$ on the SMART (BERT) model in terms of the average score. 
\m{DeBERTa (Large) and} ALBERT (XXLarge) achieve the highest average \method scores among all the tested language models. \m{This result is also aligned with the ANLI leaderboard\footnote{https://github.com/facebookresearch/anli}, which shows that ALBERT (XXLarge) is the most robust to human-crafted adversarial NLI dataset \citep{anli}. }

We note that although our adversarial examples are generated from surrogate models based on BERT and RoBERTa, these examples have high transferability between models after our data curation. Specifically, the average score of ELECTRA (Large) on \method is even lower than RoBERTa (Large), which demonstrates that \method can effectively transfer across models of different architectures and unveil the vulnerabilities shared across multiple models. 
Moreover, we find  some models even perform worse than random guess. For example, the performance of BERT on \method for all tasks is lower than  random-guess accuracy. 

We also benchmark advanced robust training methods to evaluate whether these methods can indeed provide robustness improvement on \method and to what extent.
We observe that SMART and FreeLB are particularly helpful to improve  robustness for RoBERTa.
Specifically, SMART (RoBERTa) improves RoBERTa (Large) over $3.71\%$ on average, and it even improves the benign accuracy as well. 
Since InfoBERT is not evaluated on GLUE, we run InfoBERT with different hyper-parameters and report the best accuracy on  benign GLUE dev set and \method test set. However, we find that the benign accuracy of InfoBERT (RoBERTa) is still lower than RoBERTa (Large), and similarly for the robust accuracy. 
These results suggest that existing robust training methods only have incremental robustness improvement, and there is still a long way to go to develop robust models to achieve satisfactory performance on \method.

\textbf{Diagnostic Report of Model Vulnerabilities.}
To have a systematic understanding of which adversarial attacks language models are vulnerable to, we provide a detailed diagnostic report in Table \ref{tab:report}. 
We observe that models are most vulnerable to human-crafted examples, where complex linguistic phenomena (\emph{e.g.}, numerical reasoning, negation and coreference resolution) can be found. For sentence-level perturbations, models are more vulnerable to distraction-based perturbations than directly manipulating syntactic structures. In terms of word-level perturbations, models are similarly vulnerable to different word replacement strategies, among which typo-based perturbations and knowledge-guided perturbations are the most effective attacks.

We hope the above findings can help researchers systematically examine their models against different adversarial attacks, thus also devising new methods to defend against them. Comprehensive  analysis of the model robustness report is provided in our website and Appendix \ref{appendix:website}.

\section{Conclusion}
We introduce \method, a multi-task benchmark to evaluate and analyze the  robustness of state-of-the-art language models and robust training methods. We  systematically conduct 14  adversarial attacks on GLUE tasks and adopt crowd-sourcing to guarantee the quality and validity of generated adversarial examples. 
Modern language models perform poorly on \method, 
suggesting that model vulnerabilities to adversarial attacks still remain unsolved. We hope \method can serve as a comprehensive and reliable diagnostic benchmark for researchers to further develop  robust  models.

\begin{ack}
We thank the anonymous reviewers for their constructive feedback.
We also thank Prof. Sam Bowman, Dr. Adina Williams, Nikita Nangia,  Jinfeng Li, and many others for the helpful discussion. We thank Prof. Robin Jia and Yixin Nie for allowing us to incorporate their datasets as part of the evaluation. We thank the SQuAD team for allowing us to use their website template and submission tutorials.
This work is partially supported by the NSF grant No.1910100, NSF CNS 20-46726 CAR, the Amazon Research Award.
\end{ack}

\bibliographystyle{abbrvnat}
\bibliography{main}

\clearpage

\appendix

\section{Appendix}

\subsection{Glossary of Adversarial Attacks}

\m{We present a glossary of adversarial attacks considered in \method in Table \ref{tab:glossary} and \ref{tab:glossary2}.}

\begin{table}[htp!]\small \setlength{\tabcolsep}{7pt}
\centering
\caption{\m{\small \textbf{Glossary of adversarial attacks (word-level and sentence-level) in \method.} For each adversarial attack, we provide a brief explanation and a corresponding example in \method.}
}

 \label{tab:glossary}
\resizebox{1.0\textwidth}{!}{
\begin{tabular}{p{2.0cm}p{6cm}p{7cm}}
\toprule 
 \multirow{2}{*}{\textbf{Perturbations}} &  \multirow{2}{*}{\textbf{Explanation}} & \textbf{Examples} (\st{Strikethrough} = Original Text, \textcolor{red}{red} = Adversarial Perturbation) \\
\midrule
 \multirow{6}{2.0cm}{TextBugger (Word-level / Typo-based)} & \multirow{6}{6cm}{TextBugger first identifies the important words in each sentence and then replaces them with carefully crafted typos.} & \textbf{Task:} QNLI \\
 & & \textbf{Question}: What was the population of the Dutch Republic before this emigration?  \\
 & & \textbf{Sentence}: This was a \st{huge} \textcolor{red}{hu ge} influx as the entire population of the Dutch Republic amounted to ca. \\
 & & \textbf{Prediction:} False $\rightarrow$ True \\
 \midrule
 \multirow{7}{2.0cm}{TextFooler (Word-level / Embedding-similarity-based)} & \multirow{7}{6cm}{Embedding-similarity-based adversarial attacks such as TextFooler select synonyms according to the cosine similarity of word embeddings. Words that have high similarity scores will be used as candidates to replace original words in the sentences.} & \textbf{Task:} QQP \\
 & & \textbf{Question 1}: I am getting fat on my lower body and on the \st{chest} \textcolor{red}{torso}, is there any way I can get fit without looking skinny fat?  \\
 & & \textbf{Question 2}: Why I am getting skinny instead of losing body fat? \\
 & & \textbf{Prediction:} Not Equivalent $\rightarrow$ Equivalent \\
  \midrule
 \multirow{7}{2.0cm}{BERT-ATTACK (Word-level / Context-aware)} & \multirow{7}{6cm}{BERT-ATTACK uses pre-trained BERT to perform masked language prediction to generate contextualized potential word replacements for those crucial words.} & \textbf{Task:} MNLI \\
 & & \textbf{Premise}: Do you know what this is? With a dramatic gesture she flung back the left side of her \st{coat} \textcolor{red}{sleeve}  and exposed a small enamelled badge.  \\
 & & \textbf{Hypothesis}: The coat that she wore was long enough to cover her knees . \\
 & & \textbf{Prediction:} Neutral $\rightarrow$ Contradiction \\
  \midrule
 \multirow{6}{2.0cm}{SememePSO (Word-level / Knowledge-guided)} & \multirow{6}{6cm}{Knowledge-guided adversarial attacks such as SememePSO use external knowledge base such as HowNet or WordNet to search for substitutions.} & \textbf{Task:} QQP  \\
 & & \textbf{Question 1}: What people who you've never met have \st{influenced} \textcolor{red}{infected} your life the most?  \\
 & & \textbf{Question 2}: Who are people you have never met who have had the greatest influence on your life? \\
 & & \textbf{Prediction:} Equivalent $\rightarrow$ Not Equivalent \\
  \midrule
 \multirow{5}{2.0cm}{CompAttack (Word-level / Compositions)} & \multirow{5}{6cm}{CompAttack is a whitebox-based adversarial attack that integrates all other word-level perturbation methods in one algorithm to evaluate model robustness to various adversarial transformations.} & \textbf{Task:} SST-2 \\
 & & \textbf{Sentence}: The primitive force of this film seems to \st{bubble} \textcolor{red}{bybble} up from the vast collective memory of the combatants.  \\
 & & \textbf{Prediction:} Positive $\rightarrow$ Negative \\
  \midrule
 \multirow{7}{2.0cm}{SCPN (Sent.-level / Syntactic-based)} & \multirow{7}{6cm}{SCPN is an attack method based on syntax tree transformations. It is trained to produce a paraphrase of a given sentence with specified syntactic structures.} & \textbf{Task:} RTE \\
 & & \textbf{Sentence 1}: He became a boxing referee in 1964 and became most well-known for his decision against Mike Tyson, during the Holyfield fight, when Tyson bit Holyfield's ear.  \\
 & & \textbf{Sentence 2}: Mike Tyson bit \st{Holyfield's ear} in 1964. \\
 & & \textbf{Prediction:} Not Entailment $\rightarrow$ Entailment \\
  \midrule
 \multirow{5}{2.0cm}{T3 (Sent.-level / Syntactic-based)} & \multirow{5}{6cm}{T3 is a whitebox attack algorithm that can add perturbations on different levels of the syntax tree and generate the adversarial sentence.} & \textbf{Task:} MNLI \\
 & & \textbf{Premise}: What's truly striking, though, is that Jobs \st{has} \textcolor{red}{had} never really let this idea go.  \\
 & & \textbf{Hypothesis}: Jobs never held onto an idea for long. \\
 & & \textbf{Prediction:} Contradiction $\rightarrow$ Entailment \\
  \midrule
 \multirow{5}{2.0cm}{AdvFever (Sent.-level / Syntactic-based)} & \multirow{5}{6cm}{Entailment preserving rules proposed by AdvFever transform all the sentences satisfying the templates into semantically equivalent ones.} & \textbf{Task:} SST-2 \\
 & & \textbf{Sentence}: \st{I'll bet the video game is} \textcolor{red}{There exists} a lot more fun than the film \textcolor{red}{that goes by the name of i 'll bet the video game}. \\
 & & \textbf{Prediction:} Negative $\rightarrow$ Positive \\
  \midrule
 \multirow{8}{2.0cm}{StressTest (Sent.-level / Distraction-based)} & \multirow{8}{6cm}{StressTest appends three true statements (``and true is true'', ``and false is not true'', ``and true is true'' for five times) to the end of the hypothesis sentence for NLI tasks.} & \textbf{Task:} RTE \\
 & & \textbf{Sentence 1}: Yet, we now are discovering that antibiotics are losing their effectiveness against illness. Disease-causing bacteria are mutating faster than we can come up with new antibiotics to fight the new variations.  \\
 & & \textbf{Sentence 2}: Bacteria is winning the war against antibiotics \textcolor{red}{and true is true}. \\
 & & \textbf{Prediction:} Entailment $\rightarrow$ Not Entailment \\
  \midrule
 \multirow{6}{2.0cm}{CheckList (Sent.-level / Distraction-based)} & \multirow{6}{6cm}{CheckList adds randomly generated URLs and handles to distract model attention.} & \textbf{Task:} QNLI \\
 & & \textbf{Question}: What was the population of the Dutch Republic before this emigration? \textcolor{red}{https://t.co/DlI9kw}  \\
 & & \textbf{Sentence}: This was a huge influx as the entire population of the Dutch Republic amounted to ca. \\
 & & \textbf{Prediction:} False $\rightarrow$ True \\
\bottomrule
\end{tabular}

}
\end{table}

\begin{table}[htp!]\small \setlength{\tabcolsep}{7pt}
\centering
\caption{\m{\small\textbf{Glossary of adversarial attacks (human-crafted) in \method.} For each adversarial attack, we provide a brief explanation and a corresponding example in \method.}}
 \label{tab:glossary2}
\resizebox{1.0\textwidth}{!}{
\begin{tabular}{p{2.0cm}p{6cm}p{7cm}}
\toprule 
 \multirow{2}{*}{\textbf{Perturbations}} &  \multirow{2}{*}{\textbf{Explanation}} & \textbf{Examples} (\st{Strikethrough} = Original Text, \textcolor{red}{red} = Adversarial Perturbation) \\
  \midrule
 \multirow{5}{2.0cm}{CheckList (Human-crafted)} & \multirow{5}{6cm}{CheckList analyses different capabilities of NLP models using different test types. We adopt two capability tests: \textit{Temporal} and \textit{Negation}, which test if the model understands the order of events and if the model is sensitive to negations.} & \textbf{Task:} SST-2 \\
 & & \textbf{Sentence}: I think this movie is perfect, but I used to think it was annoying. \\
 & & \textbf{Prediction:} Positive $\rightarrow$ Negative \\
 & & \\
  \midrule
 \multirow{8}{2.0cm}{StressTest (Human-crafted)} & \multirow{8}{6cm}{StressTest proposes carefully crafted rules to construct ``stress tests'' and evaluate robustness of NLI models to specific linguistic phenomena. Here we adopt the test cases focusing on \textit{Numerical Reasoning}.} & \textbf{Task:} MNLI \\
 & & \textbf{Premise}: If Anne’ s speed were doubled, they could clean their house in 3 hours working at their respective rates.  \\
 & & \textbf{Hypothesis}: If Anne’ s speed were doubled, they could clean their house in less than 6 hours working at their respective rates. \\
 & & \textbf{Prediction:} Entailment $\rightarrow$ Contradiction \\
  \midrule
 \multirow{8}{2.0cm}{ANLI (Human-crafted)} & \multirow{8}{6cm}{ANLI is a large-scale NLI dataset collected iteratively in a human-in-the-loop manner. The sentence pairs generated in each round form a comprehensive dataset that aims at examining the vulnerability of NLI models.} & \textbf{Task:} MNLI \\
 & & \textbf{Premise}: Kamila Filipcikova (born 1991) is a female Slovakian fashion model. She has modeled in fashion shows for designers such as Marc Jacobs, Chanel, Givenchy, Dolce \& Gabbana, and Sonia Rykiel. And appeared on the cover of Vogue Italia two times in a row. \\
 & & \textbf{Hypothesis}: Filipcikova lives in Italy. \\
 & & \textbf{Prediction:} Neutral $\rightarrow$ Contradiction \\
  \midrule
 \multirow{6}{2.0cm}{AdvSQuAD (Human-crafted)} & \multirow{6}{6cm}{AdvSQuAD is an adversarial dataset targeting at reading comprehension systems. Examples are generated by appending a distracting sentence to the end of the input paragraph. We adopt the distracting sentences and questions in the \textit{QNLI} format with labels ``not answered''.} & \textbf{Task:} QNLI \\
 & & \textbf{Question}: What day was the Super Bowl played on?  \\
 & & \textbf{Sentence}: The Champ Bowl was played on August 18th,1991. \\
 & & \textbf{Prediction:} False $\rightarrow$ True \\
 & & \\
\bottomrule
\end{tabular}
}
\end{table}

\subsection{Additional Related Work}

We discuss more related work about textual adversarial attacks and defenses in this subsection.

\paragraph{Textual Adversarial Attacks}
Recent research has shown deep neural networks (DNNs) are vulnerable to adversarial examples that are carefully crafted to fool machine learning models without disturbing human perception  \citep{Goodfellow2015ExplainingAH,Papernot2016DistillationAA,MoosaviDezfooli2016DeepFoolAS}. However, compared with a large amount of adversarial attacks in continuous data domain \citep{Yang2018CharacterizingAA,Carlini2018AudioAE,Eykholt2017RobustPA}, there are a few studies focusing on the discrete text domain. Most existing gradient-based attacks on image or audio models are no longer applicable to NLP models, as words are intrinsically discrete tokens. Another challenge for generating adversarial text is to ensure the semantic and syntactic coherence and consistency. 

Existing textual adversarial attacks can be roughly divided into three categories: word-level transformations, sentence-level attacks, and human-crafted samples. ($i$) Word-level transformations adopt different word replacement strategies during attack. For example, existing work \citep{textbugger, hotflip} applies character-level perturbation to carefully crafted typo words (\emph{e.g.}, from ``foolish'' to ``fo0lish''), thus making the model ignore or misunderstand the original statistical cues.  Others adopt knowledge-based perturbation and utialize knowledge base to constrain the search space. For example, \citet{comattack} uses sememe-based knowledge base from HowNet~\citep{hownet} to construct a search space for word substitution. Some \citep{textfooler, textbugger}  use non-contextualized word embedding from  GLoVe~\citep{DBLP:conf/emnlp/PenningtonSM14} or Word2Vec~\citep{DBLP:conf/nips/MikolovSCCD13} to build synonym candidates, by querying the cosine similarity or euclidean distance between the original and candidate word and selecting the closet ones as the replacements. Recent work \citep{bae,
bertattack} also leverages BERT to generate contextualized perturbations by masked language modeling. 
($ii$) Different from the dominant word-level adversarial attacks, sentence-level adversarial attacks perform sentence-level transformation or paraphrasing by perturbing the syntactic structures based on human crafted rules \citep{stresstest, checklist} or carefully designed auto-encoders \citep{scpn, t3}.  Sentence-level manipulations are generally more challenging than word-level attacks, because the perturbation space for syntactic structures are limited compared to word-level perturbation spaces that grow exponentially with the sentence length. However, sentence-level attacks tend to have higher linguistic quality than word-level, as both semantic and syntactic coherence are taken into considerations when generating adversarial sentences.
($iii$) Human-crafted adversarial examples are generally crafted in the human-in-the-loop manner \citep{advsquad,anli,dynabenchqa} or use manually crafted templates to generate test cases \citep{stresstest,checklist}. Our \method incorporates all of the above textual adversarial to provide a comprehensive and systematic diagnostic report over existing state-of-the-art large-scale language models.

\paragraph{Defenses against Textual Adversarial Attacks} To defend against textual adversarial attacks, existing work can be classified into three categories:
($i$) \textit{Adversarial Training} is a practical method to defend against adversarial examples. Existing work either uses PGD-based attacks to generate adversarial examples in the embedding space of NLP as data augmentation \citep{freelb}, or regularizes the standard objective using virtual adversarial training \citep{smart,alum,gan2020large}. However, one drawback is that the threat model is often unknown, which renders adversarial training less effective when facing unseen attacks.
($ii$) \textit{Interval Bound Propagation} (IBP) \citep{ibp} is proposed as a new technique to consider the worst-case perturbation theoretically. Recent work \citep{ibp1,ibp2} has applied IBP in the NLP domain to certify the robustness of models. However, IBP-based methods rely on strong assumptions of model architecture and are difficult to adapt to recent transformer-based language models. 
($iii$) \textit{Randomized Smoothing} \citep{DBLP:conf/icml/CohenRK19} provides a tight robustness guarantee in $\ell_2$ norm by smoothing the classifier with Gaussian noise. \citet{safer} adapts the idea to the NLP domain, and replace the Gaussian noise with synonym words to certify the robustness as long as adversarial word substitution falls into predefined synonym sets. However, to guarantee the completeness of the synonym set is challenging.

\subsection{Task Descriptions, Statistics and Evaluation Metrics}
\label{appendix:tasks}

We present the detailed label distribution statistics and evaluation metrics of GLUE and \method benchmark in \ref{tab:labeldist}.

\paragraph{SST-2} The Stanford Sentiment Treebank \cite{socher2013recursive} consists of sentences from movie reviews and human annotations of their sentiment. Given a review sentence, the task is to predict the sentiment of it. Sentiments can be divided into two classes: positive and negative.

\paragraph{QQP} The Quora Question Pairs (QQP) dataset is a collection of question pairs from the community question-answering website Quora. The task is to determine whether a pair of questions are semantically equivalent.

\paragraph{MNLI} The Multi-Genre Natural Language Inference Corpus \cite{williams2017broad} consists of sentence pairs with textual entailment annotations. Given a premise sentence and a hypothesis sentence, the task is to predict whether the premise entails the hypothesis (entailment), contradicts the hypothesis (contradiction), or neither (neutral)

\paragraph{QNLI} Question-answering NLI (QNLI) dataset consists of question-sentence pairs modified from The Stanford Question Answering Dataset \cite{rajpurkar2016squad}. The task is to determine whether the context sentence contains the answer to the question.

\paragraph{RTE} The Recognizing Textual Entailment (RTE) dataset is a combination of a series of data from annual textual entailment challenges. Examples are constructed based on news and Wikipedia text. The task is to predict the relationship between a pair of sentences. For consistency, the relationship can be classified into two classes: entailment and not entailment, where neutral and contradiction are seen as not entailment.

\begin{table}[t]
\small
    \centering
    \caption{\small The label distribution of \method dataset. For SST-2, we report the label distribution as \textit{``negative''}:\textit{``positive''}. For QQP, we report the label distribution as \textit{``not equivalent''}:\textit{``equivalent''}. For QNLI, we report the label distribution as \textit{``true''}:\textit{``false''}. For RTE, we report the label distribution as \textit{``entailment''}:\textit{``not entailment''}. For MNLI, we report the label distribution as \textit{``entailment''}:\textit{``neutral''}:\textit{``contradiction''}.
    }
    \label{tab:labeldist}
\resizebox{1.0\textwidth}{!}
{
    \begin{tabular}{ccccccc}
    \toprule
        \textbf{Corpus} & \textbf{Task} & \shortstack{\textbf{|Dev|} \\ \scriptsize{(GLUE)}} & \shortstack{\textbf{|Test|} \\  \scriptsize{(GLUE)}} & \shortstack{\textbf{|Dev|} \\ \scriptsize{(AdvGLUE)}} & \shortstack{\textbf{|Test|} \\  \scriptsize{(AdvGLUE)}} &  \textbf{Evaluation Metrics}   \\   
        \midrule
        SST-2 & sentiment & 428:444 & 1821 & 72:76 & 590:830 & acc. \\
        QQP  & paraphrase & 25,545:14,885 & 390,965 & 46:32 & 297:125 & acc./F1 \\
        QNLI & NLI/QA & 2,702:2,761 & 5,463 & 74:74 & 394:574 & acc. \\
        RTE  & NLI & 146:131 & 3,000 & 35:46 & 123:181 & acc. \\
        MNLI & NLI & 6,942:6,252:6,453 & 19,643 & 92:84:107 & 706:565:593 & matched acc./mismatched acc. \\
        \bottomrule
        \end{tabular}
    }
\end{table}

We also show the detailed per-task model performance on \method and GLUE in Table \ref{tab:fullbenchmark}.

\begin{table}[t]
\small
    \centering
    \caption{Model performance on AdvGLUE test set and GLUE dev set.}
    \label{tab:fullbenchmark}
\resizebox{1.0\textwidth}{!}{
    \begin{tabular}{l|cccccccccccc}
    \toprule
    \multirow{2}{*}{\textbf{Models}} & \multicolumn{2}{c}{\textbf{Avg}} & \multicolumn{2}{c}{\textbf{SST-2}} &  \multicolumn{2}{c}{\textbf{MNLI}} &  \multicolumn{2}{c}{\textbf{RTE}} &  \multicolumn{2}{c}{\textbf{QNLI}} &   \multicolumn{2}{c}{\textbf{QQP}}   \\
         & \scriptsize{GLUE} & \scriptsize{AdvGLUE} & \scriptsize{GLUE} & \scriptsize{AdvGLUE} & \scriptsize{GLUE} & \scriptsize{AdvGLUE} & \scriptsize{GLUE} & \scriptsize{AdvGLUE} & \scriptsize{GLUE} & \scriptsize{AdvGLUE} & \scriptsize{GLUE} & \scriptsize{AdvGLUE} \\
    \midrule 
    {BERT(Large)} & 85.76 & 33.68 & 93.23 & 33.03 & 85.78/85.57 & 28.72/27.05 & 68.95 & 40.46 & 91.91 & 39.77 & 90.72/87.38 & 37.91/16.56 \\
    {RoBERTa(Large)} & 91.44 & 50.21 & 95.99 & 58.52 & 89.74/89.86 & 50.78/39.62 & 86.60 & 45.39 & 94.14 & 52.48 & 91.99/89.37 & 57.11/41.80 \\
    \m{{T5(Large)}} & \m{90.39} & \m{56.82} & \m{95.53} & \m{60.56} & \m{88.98/89.20} & \m{48.43/38.98} & \m{84.12} & \m{62.83} & \m{93.78} & \m{57.64} & \m{90.82/88.07} & \m{63.03/55.68} \\
    {ALBERT(XXLarge)} & 91.87 & 59.22 & 95.18 & 66.83 & 89.29/89.88 & 51.83/44.17 & 88.45 & 73.03 & 95.26 & 63.84 & 92.26/89.49 & 56.40/32.35 \\
    {ELECTRA(Large)} & 93.16 & 41.69 & 97.13 & 58.59 & 90.71 & 14.62/20.22 & 90.25 & 23.03 & 95.17 & 57.54 & 92.56 & 61.37/42.40 \\
    \m{{DeBERTa(Large)}} & \m{92.67} & \m{60.86} & \m{96.33} & \m{57.89} & \m{90.95/90.85} & \m{58.36/52.46} & \m{90.25} & \m{78.94} & \m{94.86} & \m{57.85} & \m{92.29/89.69} & \m{60.43/47.98} \\
    {SMART(BERT)} & 85.70 & 30.29 & 93.35 & 25.21 & 84.72/85.34 & 26.89/23.32 & 69.68 & 38.16 & 91.71 & 34.61 & 90.25/87.22 & 36.49/20.24 \\
    {SMART(RoBERTa)} & 92.62 & 53.71 & 96.56 & 50.92 & 90.75/90.66 & 45.56/36.07 & 90.98 & 70.39 & 95.04 & 52.17 & 91.20/88.44 & 64.22/44.28 \\
    {FreeLB(RoBERTa)} & 92.28 & 50.47 & 96.44 & 61.69 & 90.64 & 31.59/27.60 & 86.69 & 62.17 & 95.04 & 62.29 & 92.58 & 42.18/31.07 \\
    {InfoBERT(RoBERTa)} & 89.06 & 46.04 & 96.22 & 47.61 & 89.67/89.27 & 50.39/41.26 & 74.01 & 39.47 & 94.62 & 54.86 & 92.25/89.70 & 49.29/35.54 \\
    \bottomrule
    \end{tabular}
}
\end{table}

\subsection{Implementation Details of Adversarial Attacks}
\label{appendix:imple}

\paragraph{TextBugger} 
To ensure the small magnitude of the perturbation, we consider the following five strategies: ($i$) randomly inserting a space into a word; ($ii$) randomly deleting a character of a word; ($iii$) randomly replacing a character of a word with its adjacent character in the keyboard; ($iv$) randomly replacing a character of a word with its visually similar counterpart (\emph{e.g.}, ``0'' v.s. ``o'', ``1'' v.s. ``l''); and ($v$) randomly swapping two characters in a word. The first four strategies guarantee the word edit distance between the typo word and its original word to be 1, and that of the last strategy is limited to 2. Following the default setting, in Strategy ($i$), we only insert a space into a word when the word contains less than $6$ characters. In Strategy ($v$), we swap characters in a word only when the word has more than $4$ characters.

\paragraph{TextFooler}  Concretely, for the sentiment analysis tasks, we set the cosine similarity threshold to be $0.8$, which encourages the synonyms to be semantically close to original ones and enhances the quality of adversarial data. For the rest of the tasks, we follow the default hyper-parameter to set the cosine similarity threshold to be $0.7$. Besides, the number of synonyms for each word is set to $50$ following the default setting.

\paragraph{BERT-ATTACK} We follow the hyper-parameters from the official codebase, and set the number of candidate words to 48 and cosine similarity threshold to $0.4$ in order to filter out antonyms using synonym dictionaries, as BERT masked language model does not distinguish synonyms and antonyms. 

\paragraph{SememePSO} We adopt the official hyper-parameters in which maximum and minimum inertia weights are set to $0.8$ and $0.2$, respectively. We also set the maximum and minimum movement probabilities of the particles to $0.8$ and $0.2$, respectively, following the default setting. Population size is set to $60$ in every task.

\paragraph{CompAttack} We follow the T3 \citep{t3} and C\&W attack \citep{Carlini2018AudioAE} and design the same optimization objective for adversarial perturbation generation in the embedding space as:
\begin{equation}\small
    \mathcal{L}(\boldsymbol{e^*})  =  ||\boldsymbol{e^*}||_p + c \cdot g(\boldsymbol{x'}),
    \label{cw}
\end{equation}
where the first term controls the magnitude of perturbation, while $g(\cdot)$ is the attack objective function depending on the attack scenario. $c$ weighs the attack goal against attack cost. CompAttack constrains the perturbation to be close to pre-defined perturbation space, including typo space (\textit{e.g.}, TextBugger), knowledge space (\textit{e.g.,} WordNet) and contextualized embedding space (\textit{e.g., } BERT embedding clusters) to make sure the perturbation is valid. We can also see from Table \ref{tab:curation} that CompAttack overall has lower filter rate than other state-of-the-art attack methods.

\paragraph{SCPN}
We use the pre-trained SCPN models released by the official codebase. Following the default setting, we select the most frequent $10$ templates from ParaNMT-50M corpus \cite{wieting2017paranmt} to guide the generation process. We first parse sentences from GLUE dev set using Stanford CoreNLP. We used CoreNLP version 3.7.0 in our experiment, along with the Shift-Reduce Parser models.

\paragraph{T3}
We follow the hyper-parameters in the official setting where the scaling const is set to $1e4$ and the optimizing confidence is set to $0$. In each iteration, we optimize the perturbation vector for at most $100$ steps with learning rate $0.1$.

\paragraph{AdvFever}
We follow the entailment preserving rules proposed by the official implementation. We adopt all $23$ templates to transform original sentences into semantically equivalent ones. Many common sentence patterns in everyday life are included in these templates.

\label{appendix:compattack}

\subsection{Examples of \method benchmark}
\label{appendix:examples}

We show more comprehensive examples in Table \ref{tab:full_examples}. Examples are generated with different levels of perturbations and they all can successfully change the predictions of all surrogate models (BERT, RoBERTa and RoBERTa ensemble).

\begin{table}[t]\small \setlength{\tabcolsep}{7pt}
\centering
\caption{\small \textbf{Examples of \method benchmark}.}

 \label{tab:full_examples}
\resizebox{1.0\textwidth}{!}{
\begin{tabular}{p{0.7cm}p{2.0cm}p{9cm}p{1.8cm}}
\toprule 
Task & Linguistic Phenomenon & Samples (\st{Strikethrough} = Original Text, \textcolor{red}{red} = Adversarial Perturbation) & Label $\rightarrow$ Prediction \\
\midrule
 \multirow{2}{*}{SST-2} &\multirow{2}{*}{\makecell[l]{Typo \\ (Word-level)}} & \textbf{Sentence}: The primitive force of this film seems to \st{bubble} \textcolor{red}{bybble} up from the vast collective memory of the combatants. & \multirow{2}{*}{\makecell[l]{Positive \\ $\rightarrow$ Negative}} \\
  \midrule
 \multirow{2}{*}{SST-2} &\multirow{2}{*}{\makecell[l]{Context-aware \\ (Word-level)}} & \textbf{Sentence}: In execution , this clever idea is far \st{less} \textcolor{red}{smaller} funny than the original , killers from space. & \multirow{2}{*}{\makecell[l]{Negative \\ $\rightarrow$ Positive}} \\
  \midrule
 \multirow{2}{*}{SST-2} &\multirow{2}{*}{\makecell[l]{CheckList \\ (Human-crafted)}} & \textbf{Sentence}: I think this movie is perfect, but I used to think it was annoying. & \multirow{2}{*}{\makecell[l]{Positive \\ $\rightarrow$ Negative}} \\
  \midrule
 \multirow{3}{*}{QQP} &\multirow{3}{*}{\makecell[l]{Embedding \\ (Word-level)}} & \textbf{Question 1}: I am getting fat on my lower body and on the \st{chest} \textcolor{red}{torso}, is there any way I can get fit without looking skinny fat? & \multirow{3}{*}{\makecell[l]{Not Equivalent \\ $\rightarrow$ Equivalent}} \\
 & & \textbf{Question 2}: Why I am getting skinny instead of losing body fat? & \\
  \midrule
 \multirow{3}{*}{QQP} &\multirow{3}{*}{\makecell[l]{Syntactic \\ (Sent.-level)}} & \textbf{Question 1}: \st{Can I learn MMA at the age of 26?} \textcolor{red}{You can learn MMA at 24?} & \multirow{3}{*}{\makecell[l]{Not Equivalent \\ $\rightarrow$ Equivalent}} \\
 & & \textbf{Question 2}: Can I learn MMA at the age of 24? & \\
  \midrule
 \multirow{2}{*}{QQP} &\multirow{2}{*}{\makecell[l]{CheckList \\ (Human-crafted)}} & \textbf{Question 1}: Is Alfred Kennedy an analyst? & \multirow{2}{*}{\makecell[l]{Not Equivalent \\ $\rightarrow$ Equivalent}} \\
 & & \textbf{Question 2}: Is Alfred Kennedy becoming an analyst? & \\
  \midrule
 \multirow{2}{*}{MNLI} &\multirow{2}{*}{\makecell[l]{Typo \\ (Word-level)}} & \textbf{Premise}: uh-huh how about any \st{matching} \textcolor{red}{mathcing} programs & \multirow{2}{*}{\makecell[l]{Entailment $\rightarrow$ \\ Contradiction}} \\
 & & \textbf{Hypothesis}: What about matching programs? & \\
  \midrule
 \multirow{3}{*}{MNLI} &\multirow{3}{*}{\makecell[l]{Distraction \\ (Sent.-level)}} & \textbf{Premise}: You and your friends are not welcome here, said Severn. & \multirow{3}{*}{\makecell[l]{Entailment $\rightarrow$ \\ Contradiction}} \\
 & & \textbf{Hypothesis}: Severn said the people were not welcome there \textcolor{red}{and true is true}. & \\
  \midrule
 \multirow{5}{*}{MNLI} &\multirow{5}{*}{\makecell[l]{ANLI \\ (Human-crafted)}} & \textbf{Premise}: Kamila Filipcikova (born 1991) is a female Slovakian fashion model. She has modeled in fashion shows for designers such as Marc Jacobs, Chanel, Givenchy, Dolce \& Gabbana, and Sonia Rykiel. And appeared on the cover of Vogue Italia two times in a row. & \multirow{5}{*}{\makecell[l]{Neutral $\rightarrow$ \\ Contradiction}} \\
 & & \textbf{Hypothesis}: Filipcikova lives in Italy. & \\
  \midrule
 \multirow{4}{*}{QNLI} &\multirow{4}{*}{\makecell[l]{Distraction \\ (Sent.-level)}} & \textbf{Question}: What was the population of the Dutch Republic before this emigration? \textcolor{red}{https://t.co/DlI9kw} & \multirow{4}{*}{\makecell[l]{False $\rightarrow$ True}} \\
 & & \textbf{Sentence}: This was a huge influx as the entire population of the Dutch Republic amounted to ca. & \\
  \midrule
 \multirow{2}{*}{QNLI} &\multirow{2}{*}{\makecell[l]{AdvSQuAD \\ (Human-crafted)}} & \textbf{Question}: What day was the Super Bowl played on? & \multirow{2}{*}{\makecell[l]{False $\rightarrow$ True}} \\
 & & \textbf{Sentence}: The Champ Bowl was played on August 18th,1991. & \\
  \midrule
 \multirow{3}{*}{RTE} &\multirow{3}{*}{\makecell[l]{Knowledge \\ (Word-level)}} & \textbf{Sentence 1}: In Nigeria, by far the most populous country in sub-Saharan Africa, over 2.7 million people \st{are} \textcolor{red}{exist} infected with HIV. & \multirow{3}{*}{\makecell[l]{Not Entailment \\ $\rightarrow$ Entailment}} \\
 & & \textbf{Sentence 2}: 2.7 percent of the people infected with HIV live in Africa. & \\
  \midrule
 \multirow{4}{*}{RTE} &\multirow{4}{*}{\makecell[l]{Syntactic \\ (Sent.-level)}} & \textbf{Sentence 1}: He became a boxing referee in 1964 and became most well-known for his decision against Mike Tyson, during the Holyfield fight, when Tyson bit Holyfield's ear. & \multirow{4}{*}{\makecell[l]{Not Entailment \\ $\rightarrow$ Entailment}} \\
 & & \textbf{Sentence 2}: Mike Tyson bit \st{Holyfield's ear} in 1964. & \\
\bottomrule
\end{tabular}
}
\end{table}

\subsection{Fine-tuning Details of Large-Scale Language Models}
\label{appendix:train}

For all the experiments, we are using a GPU cluster with 8 V100 GPUs and 256GB memory.

\paragraph{BERT (Large)} For RTE, we train our model for $10$ epochs and for other tasks we train our model for $4$ epochs. Batch size for QNLI is set to $512$, and for other tasks it is set to $256$. Learning rates are all set to $2e-5$.

\paragraph{ELECTRA (Large)} We follow the official hyper-parameter setting to set the learning rate to $5e-5$ and set batch size to $32$. We train ELECTRA on RTE for $10$ epochs and train for $2$ epochs on other tasks. We set the weight decay rate to $0.01$ for every task.

\paragraph{RoBERTa (Large)} We train our RoBERTa for $10$ epochs with learning rate $2e-5$ on each task. The batch size for QNLI is $32$ and $64$ for other tasks.

\paragraph{\m{T5 (Large)}} \m{We train our T5 for $10$ epochs with learning rate $2e-5$ on each task. The batch size for QNLI is $32$ and $64$ for other tasks. We follow the templates in original paper to convert GLUE tasks into generation tasks.}

\paragraph{ALBERT (XXLarge)} We use the default hyper-parameters to train our ALBERT. For example, max training steps for SST-2, MNLI, QNLI, QQP, RTE, is $20935$, $10000$, $33112$, $14000$, $800$ respectively. For MNLI and QQP, batch size is set to $32$ and for other tasks batch size is set to $128$.

\paragraph{\m{DeBERTa (Large)}} \m{We use the official hyper-parameters to train our DeBERTa. For example, learning rate is set to $1e-5$ across all tasks. For MNLI and QQP, batch size is set to $64$ and for other tasks batch size is set to 32.}

\paragraph{SMART} For SMART(BERT) and SMART(RoBERTa), we use grid search to search for the best parameters and report the best performance among all trained models.

\paragraph{FreeLB (RoBERTa)} For FreeLB, we test every parameter combination provided by the official codebase and select the best parameters for our training.

\paragraph{InfoBERT (RoBERTa)} We set the batch size to $32$ and learning rate to $2e-5$ for all tasks.

\subsection{Human Evaluation Details}

\begin{table}[t]
\small
    \caption{The statistics of \method in the human training phase.}
    \label{tab:humanstat}
    \centering
{
    \begin{tabular}{cccccc}
    \toprule
        \multirow{2}{*}{\textbf{Corpus}} & \textbf{Pay Rate} & \textbf{\#/ Qualified} & \textbf{Human}  & \textbf{Human}  & \textbf{Fleiss}   \\   
        & \scriptsize{(per batch)} & \textbf{Workers} & \textbf{Acc. (Avg.)} & \textbf{Acc. (vote)} & \textbf{Kappa}\\
        \midrule
        SST-2  & \$0.4 & 70 & 89.2 & 95.0 & 0.738 \\
        MNLI  &  \$1.0 & 33 & 80.4 & 85.0 & 0.615 \\
        RTE    & \$1.0 & 66 & 85.8 & 92.0 & 0.602 \\
        QNLI  &  \$1.0 & 41 & 85.6 & 91.0 & 0.684 \\
        QQP   &  \$0.5 & 58 & 86.4 & 90.0 & 0.691  \\
        \bottomrule
        \end{tabular}
    }
\end{table}

\paragraph{Human Training} We present the pay rate and the number of qualified workers in Table \ref{tab:humanstat}. We also test our qualified workers on another non-overlapping 100 samples of the GLUE dev sets for each task. We can see that the human accuracy is comparable to \citep{muppet}, which means that most our selected annotators understand the GLUE tasks well. 

\paragraph{Human Filtering} The detailed filtering statistics of each stage is shown in Table \ref{tab:filter}. We can see that around $60-80\%$ of examples are filtered due to the low transferability and high word modification rate. Among the remaining samples, around $30-40\%$ examples are filtered due to the low human agreement rates (Human Consensus Filtering), and around $20-30\%$ are filtered due to the semantic changes which lead to the label changes (Utility Preserving Filtering).

\begin{table}[t]
\small
\centering
    \caption{Filter rates during data curation.}
    \label{tab:filter}
\resizebox{1.0\textwidth}{!}
{
    \begin{tabular}{cc|ccccc|c}
    \toprule
         \multirow{2}{*}{\textbf{Tasks}} & \multirow{2}{*}{\textbf{Metrics}} & \multicolumn{5}{c|}{Word-level Attacks} & \multirow{2}{*}{\textbf{Average}} \\
         \cmidrule(lr){3-7}
         & & SememePSO & TextFooler & TextBugger & CombAttack &  BERT-ATTACK &    \\
        \midrule
          \multirow{6}{*}{SST-2} & Transferability & 58.85 & 63.56 & 64.87 & 53.58 & 66.87 & 61.54 \\
           & Fidelity & 14.65 & 11.06 & 22.40 & 19.93 & 12.03 & 16.01 \\
           & Human Consensus & 10.53 & 10.56 & 2.27 & 9.92 & 7.09 & 8.07 \\
           & Utility Preserving & 6.68 & 5.43 & 0.51 & 3.20 & 3.82 & 3.93 \\
           & Filter Rate & 90.71 & 90.62 & 90.04 & 86.63 & 89.81 & 89.56 \\
           \midrule
           \multirow{6}{*}{MNLI} & Transferability & 44.16 & 43.15 & 42.58 & 35.08 & 41.80 & 41.36 \\
           & Fidelity & 36.57 & 45.94 & 37.71 & 38.14 & 38.60 & 39.39 \\
           & Human Consensus & 10.37 & 6.38 & 5.51 & 11.15 & 9.78 & 8.64 \\
           & Utility Preserving & 4.49 & 2.08 & 1.32 & 11.07 & 5.91 & 4.97 \\
           & Filter Rate & 95.59 & 97.55 & 87.12 & 95.45 & 96.10 & 94.36 \\
           \midrule
           \multirow{6}{*}{RTE} & Transferability & 55.32 & 67.38 & 41.96 & 54.20 & 60.94 & 55.96 \\
           & Fidelity & 19.83 & 7.79 & 42.18 & 23.17 & 14.25 & 21.44 \\
           & Human Consensus & 8.08 & 7.91 & 3.55 & 7.64 & 8.44 & 7.12 \\
           & Utility Preserving & 8.69 & 6.13 & 0.60 & 5.70 & 8.54 & 5.93 \\
           & Filter Rate & 91.93 & 89.21 & 88.29 & 90.72 & 92.16 & 90.46 \\
           \midrule
           \multirow{6}{*}{QNLI} & Transferability & 63.36 & 70.67 & 59.24 & 55.47 & 69.15 & 63.58 \\
           & Fidelity & 17.73 & 13.01 & 25.31 & 23.53 & 13.17 & 18.55 \\
           & Human Consensus & 10.06 & 9.80 & 6.84 & 9.98 & 9.36 & 9.21 \\
           & Utility Preserving & 3.48 & 2.41 & 1.50 & 4.94 & 4.10 & 3.29 \\
           & Filter Rate & 94.63 & 95.89 & 92.89 & 93.92 & 95.78 & 94.62 \\
           \midrule
           \multirow{6}{*}{QQP} & Transferability & 42.96 & 58.60 & 55.09 & 44.83 & 51.97 & 50.69 \\
           & Fidelity & 45.61 & 29.35 & 26.46 & 30.99 & 37.77 & 34.04 \\
           & Human Consensus & 4.38 & 4.69 & 5.19 & 10.08 & 3.94 & 5.66 \\
           & Utility Preserving & 3.79 & 3.86 & 3.16 & 7.93 & 4.60 & 4.67 \\
           & Filter Rate & 96.73 & 96.50 & 89.90 & 93.83 & 98.28 & 95.05 \\
           \bottomrule
    \end{tabular}
    }

\end{table}

\begin{figure}
    \centering
    \includegraphics[width=\linewidth]{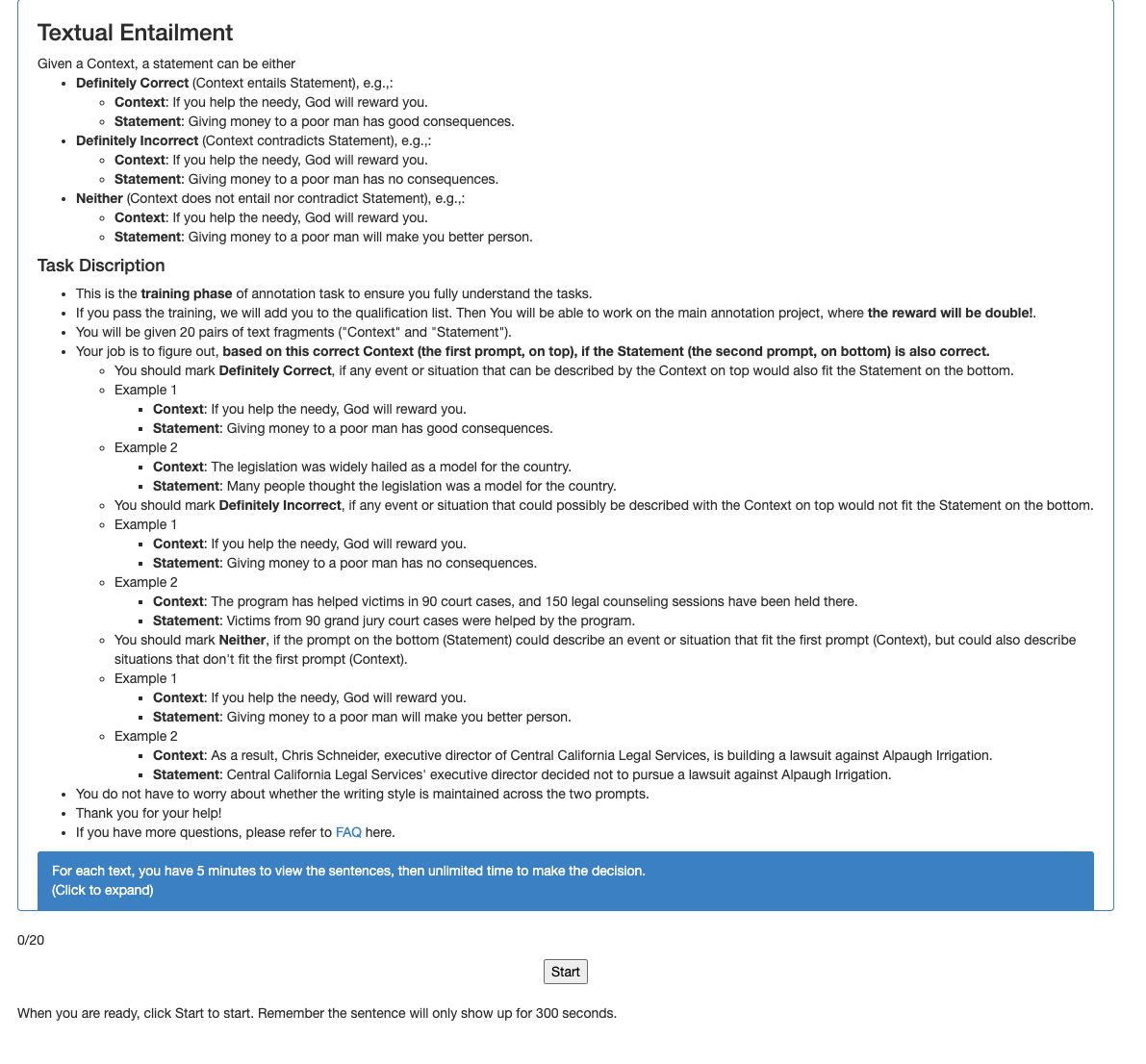}
    \caption{Human annotation instructions (training phase) for MNLI.}
    \label{fig:training}
\end{figure}

\begin{figure}
    \centering
    \includegraphics[width=\linewidth]{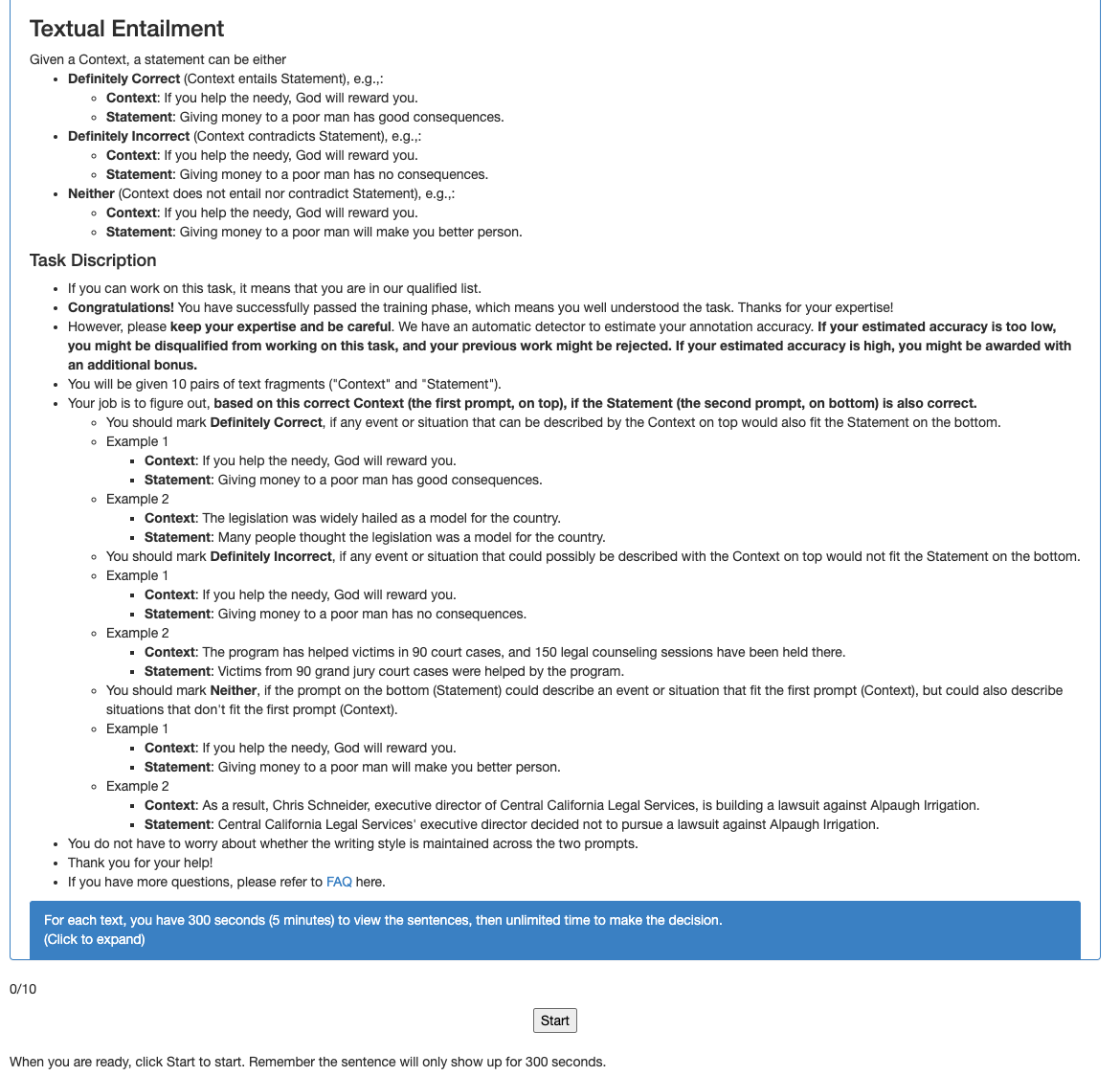}
    \caption{Human annotation instructions (filtering phase) for MNLI.}
    \label{fig:testing}
\end{figure}

\paragraph{Human Annotation Instructions} 
We show examples of annotation instructions in the training phase and filtering phase on MNLI in Figure \ref{fig:training} and \ref{fig:testing}. More instructions can be found in \url{https://adversarialglue.github.io/instructions}. We also provide a FAQ document in each task description page \url{https://docs.google.com/document/d/1MikHUdyvcsrPqE8x-N-gHaLUNAbA6-Uvy-iA5gkStoc/edit?usp=sharing}.

\subsection{Discussion of Limitations}
\label{appendix:limit}

Due to the constraints of computational resources, we are unable to conduct a comprehensive evaluation of all existing  language models. However, with the release of our leaderboard website, we are expecting researchers to actively submit their models and evaluate against our \method benchmark to have a systematic understanding of model robustness. We are also interested in the adversarial robustness of large-scale auto-regressive language models under the few-shot settings, and leave it as a compelling future work.

In this paper, we follow ANLI \citep{anli} and generate adversarial examples against surrogate models based on BERT and RoBERTa. However, there are concerns \citep{criteria} that such adversarial filtering may not be able to fairly benchmark the model robustness, as participants may top the leaderboard by producing different errors from our surrogate models. We note that such concerns can be solved given systematic data curation. As shown in our main benchmark results, we observe we successfully select the adversarial examples with high adversarial transferability that can unveil the vulnerabilities shared across models of different architectures. Specifically, we observe a huge performance gap in ELECTRA (Large) that is pre-trained with different data and shown less robust than one of surrogate model RoBERTa (Large).

Finally, we emphasize that our \method benchmark mainly focuses on robustness evaluation. Thus \method can  also be considered as a supplementary diagnostic test set besides the standard GLUE benchmark.  We suggest that participants should evaluate their models against both GLUE benchmark and our \method to understand both model generalization and robustness. We hope our work can help researchers to develop models with high generalization and adversarial robustness.

\subsection{Website}
\label{appendix:website}
\begin{figure}
    \centering
    \includegraphics[width=\linewidth,trim={0 80cm 0 0},clip]{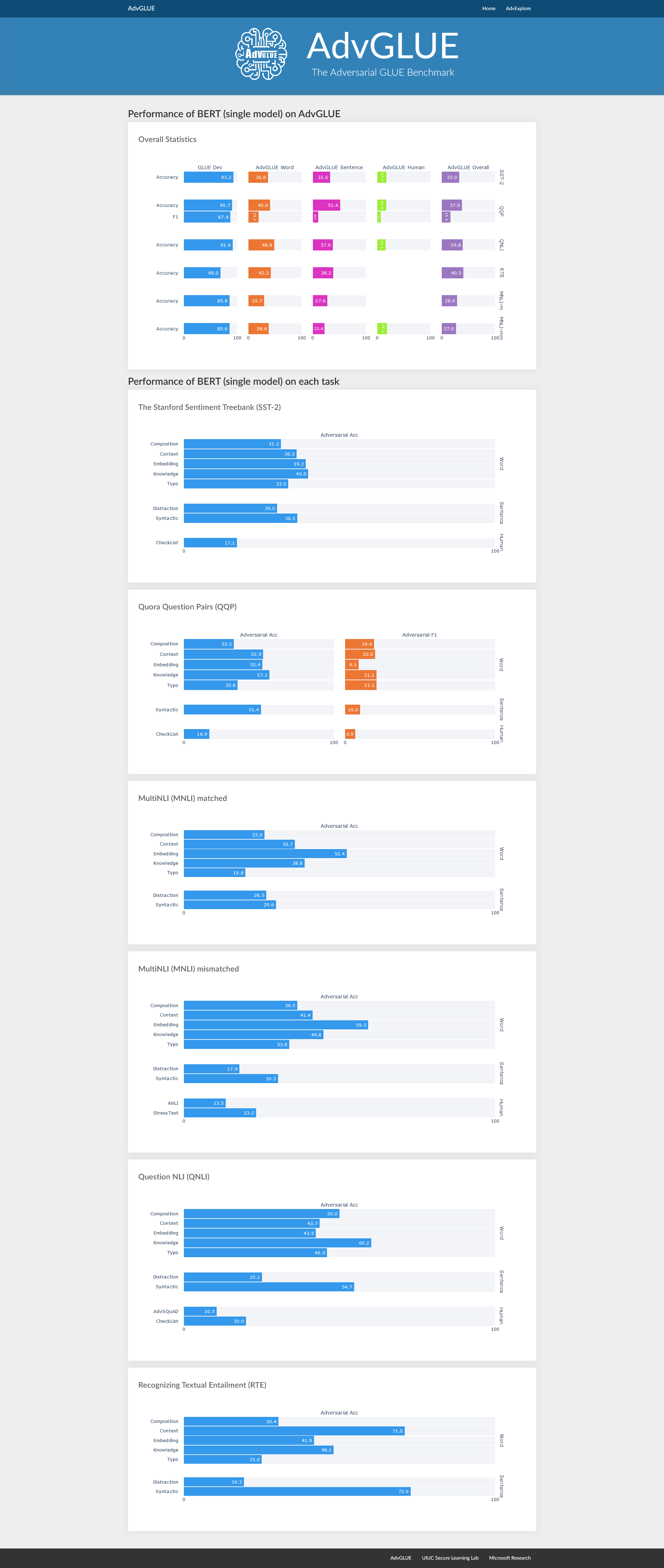}
    \caption{An example of model diagnostic report for BERT (Large).}
    \label{fig:report}
\end{figure}

We present the diagnostic report on our website in Figure \ref{fig:report}.

\section{Data Sheet}
\label{appendix:datasheet}
We follow the documentation frameworks provided by \citet{datasheet}.

\subsection{Motivation}

\paragraph{For what purpose was the dataset created?}

While recently a lot of methods (SMART, FreeLB, InfoBERT, ALUM) claim that they can improve the model robustness against adversarial attacks, the adversary setup in these methods
($i$) lacks a unified standard and is usually different across different methods;
($ii$) fails to cover comprehensive linguistic transformation (typos, synonymous substitution, paraphrasing, etc) to recognize to which levels of adversarial attacks models are still vulnerable.
This motivates us to build a unified and principled robustness benchmark dataset and evaluate to which extent the state-of-the-art models have progressed so far in terms of adversarial robustness.

\paragraph{Who created the dataset (e.g., which team, research group) and on behalf of which entity (e.g., company, institution, organization)?} University of Illinois at Urbana-Champaign (UIUC) and Microsoft Corporation.

\subsection{Composition/collection process/preprocessing/cleaning/labeling and uses:}
The answers are described in our paper as well as website \url{https://adversarialglue.github.io}.

\subsection{Distribution}

\paragraph{Will the dataset be distributed to third parties outside of the entity (e.g., company, institution, organization) on behalf of which the dataset was created?} The dev set is released to the public. The test set is hidden and can only be evaluated by an automatic submission API hosted on CodaLab.

\paragraph{How will the dataset will be distributed (e.g., tarball on website, API, GitHub)?} The dev set is released on our website \url{https://adversarialglue.github.io}. The test set is hidden and hosted on CodaLab.

\paragraph{When will the dataset be distributed?} It has been released now.

\paragraph{Will the dataset be distributed under a copyright or other intellectual property (IP) license, and/or under applicable terms of use (ToU)?} Our dataset will be distributed under the CC BY-SA 4.0 license.

\subsection{Maintenance}

\paragraph{How can the owner/curator/manager of the dataset be contacted (e.g., email address)?}

Boxin Wang (\texttt{boxinw2@illinois.edu}) and Chejian Xu (\texttt{xuchejian@zju.edu.cn}) will be responsible for maintenance.

\paragraph{Will the dataset be updated (e.g., to correct labeling errors, add new instances, delete instances)?}
Yes. If we include more tasks or find any errors, we will correct the dataset and update the leaderboard accordingly. It will be updated on our website.

\paragraph{If others want to extend/augment/build on/contribute to the dataset, is there a mechanism for them to do so?}
They can contact us via email for the contribution.

\end{document}